\documentclass[pdflatex,sn-mathphys-num]{sn-jnl}

\usepackage{graphicx}%
\usepackage{multirow}%
\usepackage{amsmath,amssymb,amsfonts}%
\usepackage{amsthm}%
\usepackage[title]{appendix}%
\usepackage{xcolor}%
\usepackage{textcomp}%
\usepackage{manyfoot}%
\usepackage{booktabs}%
\usepackage{algorithm}%
\usepackage{algorithmicx}%
\usepackage{algpseudocode}%
\usepackage{listings}%

\theoremstyle{thmstyleone}%
\theoremstyle{thmstyletwo}%

\theoremstyle{thmstylethree}%

\begin{document}

\title{Hybrid Vision-Transformer–GAN Attribute Neutralizer for Mitigating Bias in Chest X-Ray Diagnosis}


\author[1]{\fnm{Jobeal  } \sur{Solomon}}\email{jobeal.solomon@student.uva.nl}

\author*[1]{\fnm{Ali Mohammed Mansoor} \sur{Alsahag}}\email{a.m.m.alsahag@uva.nl}

\author[1]{\fnm{Seyed Sahand } \sur{Mohammadi Ziabari}}\email{s.s.mohammadiziabari@uva.nl}

\affil*[1]{\orgdiv{Informatics Institute}, \orgname{University of Amsterdam}, \orgaddress{\street{Science Park}, \city{Amsterdam}, \postcode{098XH},  \country{The Netherlands}}}


\abstract{ Bias in chest X-ray classifiers frequently stems from sex- and age-related shortcuts, leading to systematic underdiagnosis of minority sub-groups. Previous pixel-space attribute-neutralizers, which rely on convolutional encoders, lessen but do not fully remove this attribute leakage at clinically usable edit strengths. This study evaluates whether substituting the U-Net convolutional encoder with a Vision Transformer backbone in the Attribute-Neutral Framework can reduce demographic attribute leakage while preserving diagnostic accuracy. A Data-efficient Image Transformer Small (DeiT-S) neutralizer was trained on the ChestX-ray14 dataset. Its edited images, generated across eleven edit-intensity levels, were evaluated with an independent AI Judge for attribute leakage and with a convolutional neural network (ConvNet) for disease prediction. At a moderate edit level ($\alpha = 0.5$), the Vision Transformer (ViT) neutralizer reduces patient sex-recognition Area Under the Curve (AUC) to $\sim 0.80$, about 10 percentage points below the original framework’s convolutional U-Net encoder, despite being trained for only half as many epochs. Meanwhile, macro Receiver Operating Characteristic Area Under the Curve (ROC AUC) across 15 findings stays within five percentage points of the unedited baseline, and the worst-case subgroup AUC holds near 0.70. The results indicate that global self-attention vision models can further suppress attribute leakage without sacrificing clinical utility, suggesting a practical route toward fairer chest X-ray AI. }

\maketitle

\section{Introduction}
\label{sec:introduction}

Within medical image analysis, ensuring fairness in chest X-ray Artificial Intelligence (AI) is critical for mitigating health disparities, as biased algorithms can lead to underdiagnosis and undermine patient trust\cite{larrazabal2020gender, obermeyer2019dissecting}. Unfairness, characterized by uneven performance among groups identified by sensitive attributes, is often the result of AI-enabled medical systems relying on improper correlations stemming from attribute biases\cite{brown2023detecting,pagano2023bias}. If unaddressed, such bias may exacerbate diagnostic disparities and undermine trust in AI-assisted medical care.

Research conducted by Seyyed-Kalantari et al. demonstrated that diagnostic models trained on chest X-ray datasets revealed significant bias against underrepresented demographic groups, notably patients of different races, genders, and socioeconomic backgrounds, due to imbalanced or skewed training data\cite{seyyed2021underdiagnosis}. This bias translated into lower diagnostic accuracy for certain medical conditions among minorities and women, aggravating health inequities\cite{obermeyer2019dissecting}. A study conducted by Glocker et al. further highlighted how a chest X-ray model's performance in detecting certain conditions dropped by about 11\% for Black patients and 7\% for female patients, relative to the model's average performance across all subgroups\cite{glocker2023risk}. These findings highlight the need for methods that promote equitable performance across different patient attributes, specifically sex and age, which are the focus of this study.

The "Attribute-Neutral Framework," introduced by Hu et al.\cite{hu2024enhancing}, addresses this issue by neutralizing these demographic attributes. This framework mitigates unintended biases by ensuring model predictions are independent of sensitive attributes. This is achieved by the introduction of an Attribute Neutralizer that generates modified chest X-ray images in which the sensitive attribute markers are scrubbed out while preserving the medically relevant content. Training diagnostic models on these neutralized images could result in reduced model performance disparities, as the models are less able to rely on visual cues associated with sensitive patient attributes. Following attribute neutralization, an AI-Judge verifies that demographic attributes such as sex and age are no longer detectable. A disease-diagnosis model is then trained on the cleaned images. To provide a comprehensive evaluation, several fairness-enhancing algorithms are assessed at the diagnostic stage.

Research in pixel-space debiasing has explored Generative Adversarial Network (GAN)-based methods such as StarGAN and StyleEx\cite{choi2018stargan}\cite{atad2022chexplaining}. However, because these approaches use convolutional encoder-decoder architectures, it remains unclear whether pixel-level attribute neutralization generalises to Transformer backbones. 
Notably, Transformer-based models have achieved state-of-the-art performance on radiology benchmarks\cite{queiroz2025fair}. Vision Transformers, in particular, aggregate long-range cues via global self-attention and often outperform Convolutional Neural Networks (CNNs) on medical-imaging benchmarks\cite{li2023transforming}. Given Hu et al.’s finding that their GAN-based neutralizer is model-agnostic across CNN variants\cite{hu2024enhancing}, it remains unclear whether any additional leakage reduction from global self-attention can be achieved without significantly compromising clinical diagnostic performance.

To address this gap, we replaced the convolutional U-Net encoder in the Attribute-Neutral Framework with a Data-efficient Image Transformer Small (DeiT-S)\cite{touvron2021training}, disabling skip connections to isolate the impact of self-attention on fairness and utility. The resulting neutralizer therefore has a Vision Transformer encoder and the original CNN decoder. We then applied this ViT-based neutralizer to the Chest X-ray14 dataset, evaluating both attribute-recognition leakage and downstream disease-diagnosis performance across sex- and age-based subgroups. These findings, alongside insights from fairness-perception literature, can both facilitate future extensions of the Attribute-Neutral Framework to broader medical-imaging tasks and inform the design of more socially acceptable AI systems.

This study implements the Attribute-Neutral Framework to a large public chest X-ray dataset and replaces the original U-Net encoder with a ViT. The experimental setup enables an assessment of how the ViT-based neutralizer influences attribute leakage and diagnostic performance, and facilitates future extensions of the Attribute-Neutral Framework to broader medical imaging tasks. Finally, the study reflects on how these findings, together with concepts from fairness-perception literature, can inform the design of more socially acceptable AI systems for medical imaging.

The research question that will be answered during this research is as follows:

\textit{What is the impact of extending the Attribute-Neutral Framework with a Vision-Transformer on fairness and diagnostic performance in chest X-ray diagnosis?}

The following research question can be further divided into the following sub-questions:
\begin{itemize}   
    \item \textit{How does swapping the U-Net convolutional encoder for a Vision-Transformer encoder affect attribute leakage and diagnostic performance across sex- and age-based sub-groups?}    
    \item \textit{How effective is the Vision Transformer–extended Attribute Neutral Framework at reducing demographic attribute leakage while preserving diagnostic accuracy in chest X-ray AI?}
    \item \textit{What practical and stakeholder-related factors should be considered when designing chest-X-ray AI systems that are perceived as fair and clinically acceptable?}
\end{itemize}

\section{Related Work}
\label{sec:related_work}

This section reviews four key areas: pixel space attribute neutralization and image editing methods, CNN versus Transformer architectures, social equity and fairness perceptions, and dataset- and population-level fairness considerations.

\subsection{Attribute-Neutralization and Image-Editing Debiasing Methods}

Early work on debiasing hid sensitive cues inside the latent representation via adversarial training, leaving pixels intact. More recent methods enable fairness through visible edits to the input image, resulting in interpretable counterfactuals\cite{sattigeri2019fairness}. 
Fader Networks slide attribute values in latent space\cite{lample2017fader}, StarGAN unifies multi-domain translations in a single model\cite{choi2018stargan}, and AttGAN adds explicit reconstruction and attribute-classification losses, forming the backbone of Hu et al.’s Attribute-Neutral Framework\cite{he2019attgan,hu2024enhancing}.

Pixel-level editing is now common in medical imaging. StyleEx manipulates StyleGAN latents to probe chestX-ray classifiers \cite{atad2022chexplaining}, GANSan learns a Cycle-GAN-like “sanitiser” that removes demographic attributes before downstream use \cite{aivodji2021local}, and FastDiME applies diffusion-based in-painting to strip shortcut features in Chest X-ray data\cite{weng2024fast}.

In this study, Hu et al's convolutional decoder is maintained\cite{hu2024enhancing}, while the original U-Net encoder is substituted with a DeiT-S\cite{touvron2021training}, selected for its balance between predictive performance and computational efficiency. Skip connections are disabled. The modification evaluates whether introducing a self-attention backbone alters the fairness–utility trade-off. This comparison therefore sets the stage for the central test whether replacing the U-Net with a ViT can deliver fairer images without compromising diagnostic performance.

\subsection{CNN vs. Transformer Models in Medical Imaging}
Stated as limitation by Hu et al\cite{hu2024enhancing}, the effectiveness of the Attribute Neutralizer requires validation on other types of imaging models. CNNs are widely recognized for their effectiveness in X-ray diagnosis, achieving high accuracy on large-scale datasets\cite{regmi2023vision}. However, studies have revealed that CNN‑based classifiers often show significant bias across demographic groups. For instance, a CNN trained on chest X‑rays can have true positive rate disparities across sex, age, and race groups despite high overall accuracy\cite{seyyed2020chexclusion}. This indicates that high performance alone doesn’t guarantee fairness, a model may systematically under-diagnose certain groups. 

Transformer‑based models, especially Vision Transformers, are emerging as alternatives with different properties that may affect bias. ViTs rely on self‑attention instead of localized convolutions, giving them a weaker inductive bias but the ability to capture global image context. This flexibility lets ViTs scale effectively with data. When trained on large and diverse datasets, transformers can learn a broader range of features and potentially generalize better across patient subpopulations\cite{li2023transforming}. 

Prior studies have shown that transformers achieve compelling performance in tasks like disease classification, segmentation, and image reconstruction\cite{henry2022visiontransformersmedicalimaging,he2023transformers}. Results indicate that while CNNs tend to perform better when trained from scratch, readily available ViTs, especially when pretrained with ImageNet or via self-supervision, can match or even surpass CNN performance on medical imaging tasks\cite{takahashi2024comparison,matsoukas2021time}.

Beyond discriminative tasks, fully transformer-based are now on par with convolutional ones. ViTGAN replaces both generator and discriminator with ViT blocks and, with stability tricks such as improved spectral normalisation and overlapping patches, achieves image-generation quality comparable to StyleGAN-2\cite{lee2021vitgan}. In our study, we took a lighter-weight route by swapping only the Attribute Neutral Framework's U-Net encoder for a DeiT-S transformer while keeping the convolutional decoder, isolating the encoder’s impact on attribute leakage.

These trends justified comparing CNN- and ViT-based neutralizers within the Attribute-Neutral Framework, addressing Hu et al.’s limitation and informing the design of more equitable imaging AI\cite{hu2024enhancing}. Understanding such architectural drivers is essential for judging whether any fairness gain truly stems from the ViT backbone.

\subsection{Mixup and Sampling-Based Fairness Methods}
Mixup diversifies the training distribution with convex combinations of samples. FairMixup and its Manifold variant extend this idea to interpolate specifically between demographic groups, improving demographic-parity and equalised-odds scores on vision tasks\cite{chuang2021fair}. Balanced sampling mitigates bias by down-weighting over-represented subgroups. Larrazabal et al. show that training with gender-imbalanced chest X-ray data leads to a significant decrease in Area Under the Curve (AUC) for the under-represented gender, even at moderate imbalance ratios of 25\%/75\%, compared with a perfectly balanced 50\%/50\% split\cite{larrazabal2020gender}. In this study, we benchmark the ViT neutralizer’s fairness–utility trade-off against these alternative debiasing methods to ensure that any advantage is weighed against widely used pre- and in-training debias strategies.

\subsection{Social Equity and Fairness Perceptions in AI Medical-Imaging}
AI fairness in medical imaging should be grounded in social equity and medical ethics. In healthcare, unfair model outcomes violate the principle of justice by providing unequal care across patient groups\cite{liu2023translational}. Fairness frameworks should therefore incorporate social equity perspectives. Bias mitigation requires participatory, context-specific definitions of fairness beyond raw performance metrics\cite{ricci2022addressing}. Hu et al.’s work on an attribute-neutral model is effective in reducing performance gaps between groups\cite{hu2024enhancing}, but it remains unclear how image neutralization affects patient or clinician perceptions of fairness\cite{rajkomar2018ensuring}. It is essential to assess whether the removal of demographic cues affects patients’ sense of identity, autonomy, and trust in their care providers. Explainable chest X-ray systems achieve agreement with radiologists in only about half of cases, highlighting that transparency alone does not guarantee trust\cite{rong2022user}. Addressing these social dimensions is therefore essential for ensuring that AI-enabled imaging systems are both technically fair and socially acceptable. These considerations inform the design criteria for AI medical-imaging systems that clinicians and patients deem socially acceptable and trustworthy.

\subsection{Fairness Across Datasets and Populations}
Dataset composition significantly impacts model bias. Seyyed-Kalantari et al. showed that chest X-ray classifiers trained on the datasets ChestX-ray14, MIMIC-CXR or CheXpert display true-positive-rate gaps for sex, age and race, and that no single source is bias-free\cite{seyyed2020chexclusion}. Training on a multi-source mixture of the same three datasets halved those gaps, confirming that demographic diversity, not sheer size, is key to fairness. Future work should include chest X-ray datasets from other regions such as the Spanish PadChest in order to further diversify the training data\cite{bustos2020padchest}. In this study, we restrict experimentation to ChestX-ray14 due to computational budget constraints, but we retain Seyyed-Kalantari et al.'s finding as motivation for future cross-site evaluation of the ViT-based neutralizer.

\renewcommand{\arraystretch}{1.2}

\section{Methodology}
\label{sec:methodology}

This section provides an overview of the dataset, introduces the two neutralizers, outlines the generation process, details the evaluation metrics and implementation, and concludes with a discussion of built-in limitations. Due to Graphics Processing Unit (GPU) quota limits, only Chest X-ray14 was run through the full pipeline. The detailed rationale appears in Appendix\ref{app:scope}.

\subsection{Data and Pre-processing}

\subsubsection{Dataset}
\begin{figure}[htbp]
  \centering
  \includegraphics[width=\textwidth]{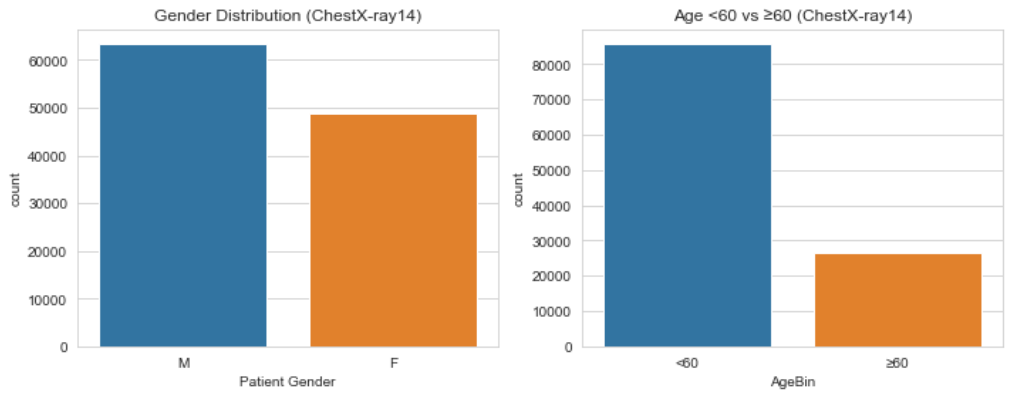}
  \caption{
  Joint distribution of sex and age bins.
  Young males dominate, whereas females aged $\ge 60$ form the smallest
  subgroup (Wilson 95 \% CI for their proportion excludes 10 \%),
  motivating the up-weighting strategy used during balanced sampling.
}
  \label{fig:cxr14_demographics}
\end{figure}

The United States-based ChestX-ray14 (NIH, 1992-2015) is the dataset carried through the full pipeline\cite{Wang_2017}. This dataset was selected for its relatively small download size, well-structured metadata in the Comma-Separated Values (CSV) file, and extensive use in medical-imaging studies. After filtering for frontal chest radiographs (posterior–anterior and anterior–posterior views, PA/AP) and removing cases with missing age or non-binary sex, the dataset contains 112\,120 images from 30\,805 patients, each annotated for 15 findings. The demographic split is 57\,\% male vs.\ 43\,\% female and 24\,\% $\geq$\,60\,y vs.\ 76\,\% $<$\,60\,y.The joint sex–age distribution (Figure~\ref{fig:cxr14_demographics}) shows that older women form the smallest subgroup, which is relevant for assessing worst-case subgroup performance. Label prevalence is highly unbalanced, ranging from 42.7\,\% (No Finding) to 0.16\,\% (Hernia).

\begin{figure}[htbp]
  \centering
  \includegraphics[width=\textwidth]{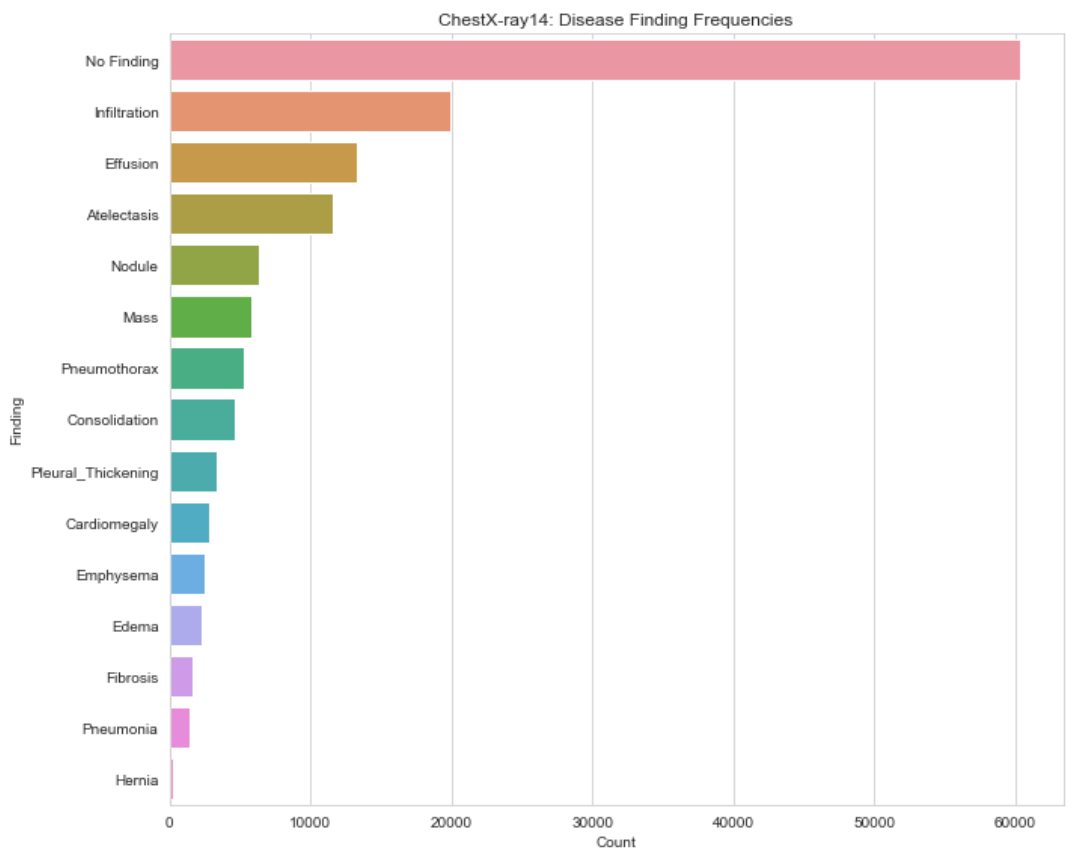}
  \caption{
  Frequencies of the 15 diagnostic labels in ChestX-ray14.
  The most common class (“No Finding’’) is over three orders of magnitude
  more frequent than the rarest (“Hernia’’), illustrating the pronounced
  long-tailed label distribution that the loss re-weighting scheme must
  address.
}
  \label{fig:cxr14_disease_freq}
\end{figure}

Figure~\ref{fig:cxr14_disease_freq} visualises this long-tailed label distribution, with “No Finding” dominating rare diagnoses such as Hernia.

\clearpage

\clearpage

\subsubsection{Pre-processing}
Retained images are converted to single‐channel PNG or JPG, resized to \(256 \times 256\), and min–max scaled to \([-1,1]\). Age is binned into two groups (\(<60\)y, \(\ge60\)y). The 15 findings are one‐hot encoded, with blank or “uncertain” labels set to 0. Data are split at the patient level into 80\,\% train, 10\,\% validation, and 10\,\% test subsets; the same split is used for every experiment.

\subsection{Model Pipeline}
\begin{figure}[htbp]
  \centering
  \hspace*{-0.5cm}%
  \includegraphics[width=0.40\textwidth]{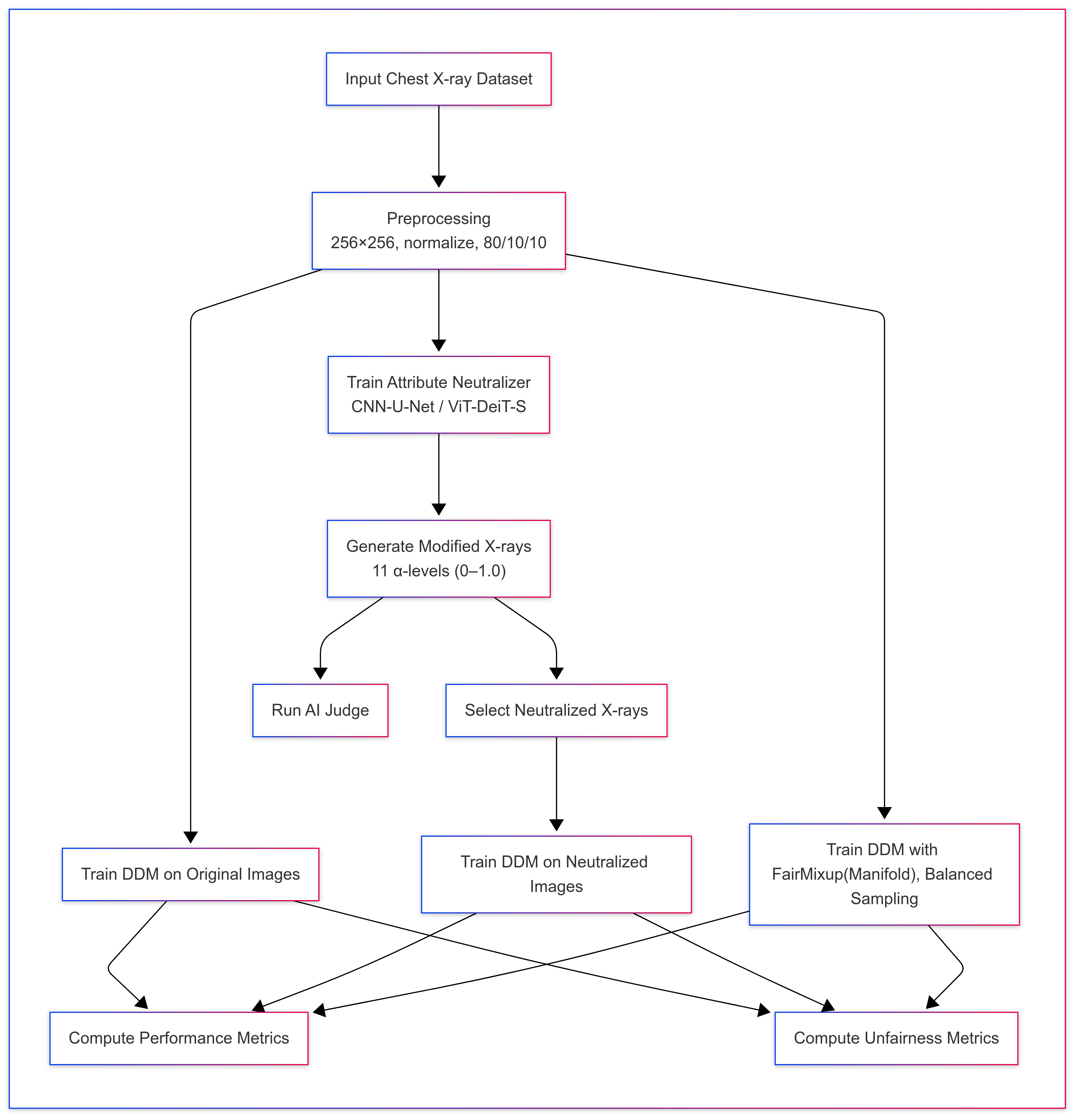}
  \caption{End-to-end pipeline: ChestX-ray14 images are pre-processed, edited by two neutralizers at 11 $\alpha$ levels, and then (i) scored for attribute leakage by an AI-Judge and (ii) used to train a Disease-Diagnosis Model (DDM). The DDM trained on original images plus two alternative debiasing baselines serve as comparators. All models are evaluated on a common test set for accuracy and fairness.}
  \label{fig:pipeline}
\end{figure}

See Figure~\ref{fig:pipeline} for the overview of the Attribute-Neutral Framework pipeline.

\subsection{Attribute Neutralizer}

\begin{figure}[htbp]
  \centering
  \includegraphics[width=\columnwidth]{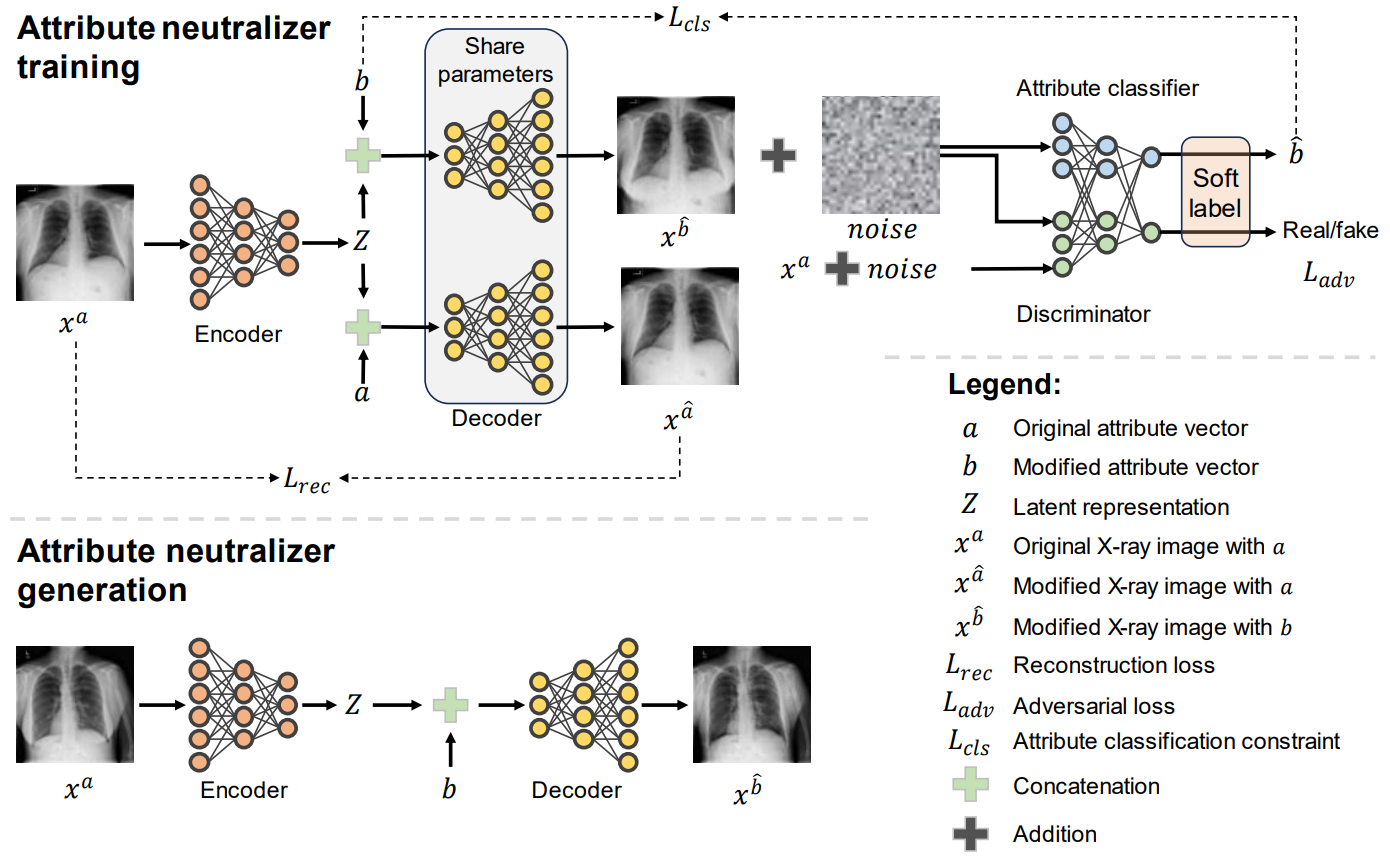}
  \caption{The Attribute Neutralizer training and image-generation process\cite{hu2024enhancing}.}
  \label{fig:attr-pipeline}
\end{figure}

The Attribute Neutralizer is designed to neutralize sensitive attributes from X-ray images, ensuring that AI models trained on these images do not rely on biased demographic cues for prediction\cite{hu2024enhancing}. In our study, we kept the original AttGAN-based design to stay aligned with Hu et al.’s framework, testing whether the pipeline, when augmented with a ViT encoder, functions effectively with an alternative backbone architecture. The Attribute Neutralizer is based on the AttGAN framework and utilizes a U-net architecture, combining a generator and a discriminator to produce neutralized images with minimal loss of medical content\cite{he2019attgan}. The process involved encoding the original image into a latent space and then modifying its attributes by decoding it back with the adjusted attribute vector. The key steps and components are as follows:

\subsubsection{Generator}
The generator comprised an encoder–decoder structure that processed the original X-ray image. The encoder captured the image’s essential features and the attribute vector, while the decoder reconstructed the modified image, with attributes adjusted according to the vector.

\subsubsection{Discriminator}
The discriminator functioned as a multi-task classifier. It distinguished between the original and modified images while simultaneously ensuring that the modified image corresponded to the desired attribute changes. The model classified whether the image was real or fake, based on whether the modified attributes aligned with the expected changes.

\subsubsection{Loss Functions}
The model is optimized using a combination of three loss functions:

\begin{itemize}
    \item \textbf{Reconstruction Loss} ($L_{\text{rec}}$): Ensures that the modified image retains the original medical information.
    \item \textbf{Attribute Classification Loss} ($L_{\text{cls}}$): Guarantees that the modified images correspond to the new attribute vector.
    \item \textbf{Adversarial Loss} ($L_{\text{adv}}$): Ensures that the modified images are indistinguishable from genuine images by the discriminator.
\end{itemize}

The loss function for the generator is given as:
\[
L_{\text{gen}} = \lambda_1 L_{\text{rec}} + \lambda_2 L_{\text{cls}} + L_{\text{adv}}
\]
For the discriminator, the loss is defined as:
\[
L_{\text{dis}} = \lambda_3 L_{\text{cls}} + L_{\text{adv}}
\]
These loss terms are weighted by hyperparameters $\lambda_1$, $\lambda_2$, and $\lambda_3$, which balance the contributions of each objective.

\subsubsection{Attribute Vector}
The sensitive attributes were encoded as binary or one-hot vectors. These vectors were manipulated to neutralize the identified attributes. A key feature of the Attribute Neutralizer was its ability to adjust multiple attributes simultaneously without requiring separate models for each. The degree of modification was controlled by a parameter $\alpha$, where 0 represented no modification, 1 represented full negation, and 0.5 signified neutralization.

\subsubsection{Training and Generation}
During training, the model learned to modify the images to reduce the predictability of sensitive attributes while retaining diagnostic accuracy. The trained model could then generate neutralized X-ray images by encoding the original image, adjusting the attribute vector, and decoding it to produce the final output. The training process also included regularization techniques such as adding Gaussian noise, label smoothing, and random image augmentations to enhance the stability of the model.

\subsubsection{Hyperparameters}
Model: U-net generator + multi-task discriminator (based on AttGAN\cite{he2019attgan}); Gaussian noise (mean = 0.1) added to discriminator input; fake/real label flip rate: 5\%; label smoothing on attribute vectors; data augmentation: random horizontal flips; convolution kernel size: $6\times6$; loss weights: reconstruction $\lambda_{1}=100$, classification $\lambda_{2}=10$, adversarial = 1, gradient penalty = 10; learning rate: $1\times10^{-4}$; batch size: 64; epochs: 300; hardware: Snellius High-Performance Computing (HPC) system.

\subsection{Vision Transformer Adaptation}
To the best of my knowledge (literature checked through June 2025), no study has yet combined a ViT-encoder with an AttGAN-based Attribute-Neutralizer for attribute-controlled editing of medical X-ray images.

 \begin{figure}[htbp]
  \centering
  \hspace*{-0.5cm}%
  \includegraphics[width=0.40\textwidth]{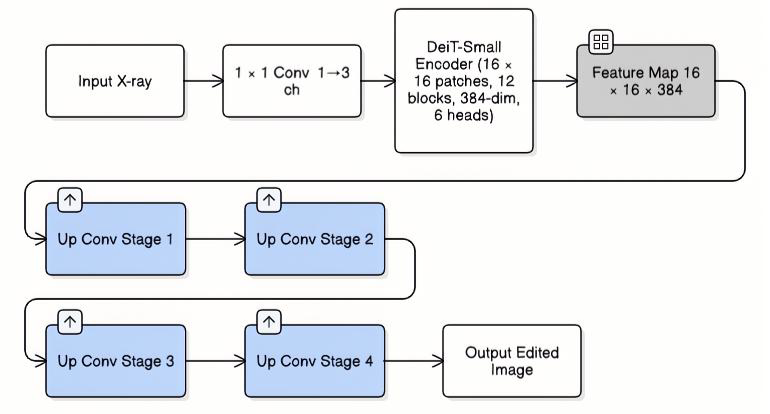}
  \caption{ViT-adapted generator. A learnable 1 × 1 convolution maps the single-channel X-ray to Red, Green, and Blue (RGB). A DeiT-Small transformer encoder (12 blocks, 16 × 16 patches, 384-dim) replaces the original CNN encoder, skip connections are removed, and the decoder depth is reduced to four up-convolution stages.}
  \label{fig:vit_adapt}
\end{figure}

\subsubsection{Architecture and model selection}
The Attribute-Neutralizer’s U-Net encoder was replaced by a DeiT-S, chosen for its favourable accuracy-to-compute trade-off in medical-imaging studies\cite{matsoukas2021time}. Each \(256\times256\) single-channel X-ray was converted to RGB through a learnable \(1\times1\) convolution so that ImageNet pre-trained weights could be reused. The image was split into \(16\times16\) patches (256 tokens), projected to a 384-dimensional embedding, and processed by 12 transformer blocks with 6-head self-attention. After the class token was discarded, the sequence was reshaped into a \(16\times16\) feature map and forwarded to the unchanged decoder. Skip connections were disabled, and the decoder depth was reduced to four up-convolutions to match the coarser latent grid. Discriminator architecture and loss weights remained unchanged at \(\lambda = 100:10:10:10\) (reconstruction, attribute, adversarial, gradient-penalty).

\subsubsection{Training schedule}
Two neutralizer runs, one for sex and one for age, were trained with identical settings: batch size 64, 150 epochs (1 750 iterations·epoch), AdamW optimiser (learning rate \(1\times10^{-4}\), \(\beta=(0.9,0.999)\), weight-decay = 0.01) and cosine learning-rate decay with linear warm-up over the first epoch. The random seed was fixed to 0 for reproducibility. The final epoch-150 checkpoints were used for all subsequent image-generation experiments.

\subsubsection{Downstream consistency}
Apart from the encoder swap, every later-stage AI-Judge leakage test, DDM training, and statistical evaluation followed the same procedure described for the CNN baseline.

\subsection{Inference and Image Generation}
Once both neutralizers were trained, their generators were run in inference-only mode to create the material for evaluation. For every X-ray in the train, validation, and test splits, 11 edited variants were generated, corresponding to the attribute edit-degree parameter \(\alpha = 0.0, 0.1, \dots, 1.0\). The edit-degree \(\alpha\) blended the original attribute embedding with its inverse embedding. When \(\alpha = 0.0\), the model output the original image; when \(\alpha = 1.0\), it swapped the protected attribute completely (e.g., female \(\rightarrow\) male). When \(\alpha = 0.5\), it generated an image that was maximally neutral with respect to that attribute. Each set of images was written as a NumPy array. The generator architecture flags were kept identical to the training run so that the stored checkpoint loaded without shape mismatches.

\subsection{AI Judge for Attribute Recognition}

The AI Judge, designed to identify the attributes of X-ray images processed by the Attribute-Neutral Framework\cite{hu2024enhancing}, was trained to recognize binary attributes in both the original and modified X-ray images. In our study, we maintained Hu et al.’s ConvNet Judge\cite{hu2024enhancing} to keep our implementation aligned with the original framework. The deep-learning model ConvNet\cite{liu2022convnet} was utilized for training. This model was pre-trained on the ImageNet dataset and fine-tuned on the X-ray images with specific data augmentation techniques\cite{hu2021splicing}, including random horizontal flips, rotations, Gaussian blur, and affine transformations. The AI Judge had two output nodes, one for each binary classification task.

\subsubsection{Hyperparameters}
Model: ConvNet32 (pre-trained on ImageNet), 2 output nodes; data augmentation: random horizontal flip, rotation, Gaussian blur, affine; learning rate: $5\times10^{-4}$; batch size: 120; epochs: 100; hardware: Snellius HPC system.

\subsection{Disease Diagnosis Model}

The Disease Diagnosis Model was tasked with identifying medical findings in the X-ray images. Likewise, ConvNet was selected for the disease diagnosis tasks. The model handled a multi-label classification task, where each image could have multiple medical findings or be classified as “No finding.” Hu et al.’s disease diagnosis model was maintained to ensure full alignment with the original framework\cite{hu2024enhancing}. The model architecture included a sigmoid activation function in the output layer, and the binary cross-entropy loss function was used to compute the loss between predicted and true labels.

\subsubsection{Hyperparameters}
Model: ConvNet (multi-label), sigmoid outputs, binary cross-entropy with class weights; initialization \& augmentation; learning rate: $5\times10^{-4}$; batch size: 120; epochs: 100, plus 20 additional epochs for ensemble evaluation; evaluation: saved model each epoch, final performance averaged over last 20; hardware for Disease Diagnosis Model: Snellius HPC system.

\subsection{Alternative Unfairness Mitigation Algorithms}
The literature review identified FairMixup/Manifold Mixup and balanced subgroup sampling as commonly adopted debiasing baselines, so we benchmarked against both. These algorithms offered diverse strategies to mitigate unfairness, though they required integration into the Disease Diagnosis Model.

\subsubsection{FairMixup}
Mixup can be implemented at two levels, with FairMixup at the image level and its Manifold variant at the feature level. FairMixup generated interpolated samples by blending images from different demographic groups, thereby smoothing the decision boundary across subgroups. FairMixup manifold applied the same interpolation in the learned feature space, producing mixed representations that promoted fairness in binary classification tasks.

\subsubsection{Balanced Sampling}
This technique mitigated unfairness by balancing the dataset across groups. It achieved this by downsampling the majority class to match the size of the minority class, ensuring proportional distribution among the various medical findings.

\subsubsection{Hyperparameters for Alternative Unfairness Mitigation Algorithms}

\textbf{FairMixup and FairMixup manifold} (baseline DDM settings): adjusted hyperparameters—batch size: 32; epochs: 50; training iterations: 300; test iterations: 150; hardware: Google Colab.  
\textbf{Balanced sampling} (baseline DDM settings): adjusted hyperparameters—epochs: 75; training iterations: 720; test iterations: 100; hardware: Google Colab.

\subsection{Statistical analysis}
\label{sec:stats}

For the image-level analysis, the Structural Similarity Index Measure (SSIM) is computed patch-wise (100$\times$100 pixels) between each neutralized image and its original counterpart, and then averaged over all patches and test images for each edit-intensity $\alpha$. Pearson's correlation coefficient $r$ is calculated between SSIM and $\alpha$, as well as between AI-Judge accuracy and $\alpha$, to confirm that stronger edits increase visual deviation while reducing attribute recognisability.

Per-model performance is evaluated using pair-wise differences in ROC\textendash AUC (effect size \(\Delta\)AUC), assessed with the two-sided DeLong test. The point estimate \(\Delta\)AUC is reported together with its 95\% confidence interval and the Benjamini\textendash Hochberg–corrected \(p\)-value across all 15 findings. For PR\textendash AUC, accuracy, sensitivity, specificity, and \(F_{1}\), the point estimate and corresponding 95\% confidence intervals are provided. As no overlap-based statistical test applies to these metrics, only descriptive effect sizes are reported. For the CNN–versus–ViT macro diagnostic performance comparison, \(\Delta\), its 95\,\% confidence interval, and a two-sided \(p\)-value are derived from 1\,000 paired image-level bootstrap resamples of the shared test subset using a sign test around zero.

To compare alternative unfairness-mitigation algorithms, a Friedman test is applied to the performance of all six methods on each protected attribute using the held-out test set. When significance is detected, a Nemenyi post-hoc rank comparison is performed and visualized using Critical-Difference diagrams. Fairness is summarized using three complementary metrics: worst-case subgroup performance (min-AUC), best-minus-worst performance gap, and the standard deviation of subgroup AUCs. All metrics are computed from the same per-finding predictions used in the global performance analysis, ensuring that each test image is evaluated only once.

\section{Results}
\label{sec:results}

\subsection{Similarity between modified and original X-ray images}
\label{sec:ssim}

Bidirectional SSIM curves for the CNN and ViT neutralizers are shown in
Figure 4, and the linear correlations with the
edit-intensity parameter~$\alpha$ are summarised in Table 1.
For the CNN encoder, SSIM declines marginally in the
male\,$\to$\,female and $<60\to\ge60$ directions, but increases
slightly in the reverse directions, indicating that the generator leaves
most pixels unchanged when converting female\,$\to$\,male or
$\ge60\to<60$.
By contrast, the ViT encoder produces larger
pixel-level edits: SSIM falls by 6–13\,\% at $\alpha=1.0$ and shows a strong negative correlation with $\alpha$ in three of the four
directions ($|r| \geq 0.96$, $p<2\times10^{-6}$), reducing the directional
asymmetry noted for the CNN model.

\begin{table}[htbp]

  \centering
  \scriptsize
  \setlength{\tabcolsep}{2pt} 
  \caption{Pearson correlation ($r$) between SSIM and $\alpha$ for 1\,000 randomly-sampled test images.}
  \label{tab:pearson_ssim_alpha}
  \begin{tabular*}{\columnwidth}{@{\extracolsep{\fill}} ll rr rr}
    \toprule
            &                      & \multicolumn{2}{c}{U-Net encoder (CNN)} & \multicolumn{2}{c}{DeiT-S (ViT) encoder} \\
    Attribute & Edit direction      & $r$     & $p$                   & $r$     & $p$ \\
    \midrule
    Gender   & male $\to$ female   & $-0.993$ & $1.5{\times}10^{-9}$  & $-0.966$ & $1.4{\times}10^{-6}$ \\
             & female $\to$ male   & $+0.990$ & $5.7{\times}10^{-9}$  & $-0.960$ & $2.9{\times}10^{-6}$ \\[2pt]
    Age      & $\ge\!60\to<\!60$   & $+0.999$ & $8.3{\times}10^{-15}$ & $-0.017$ & $9.6{\times}10^{-1}$ \\
             & $<\!60\to\ge\!60$   & $-0.999$ & $7.6{\times}10^{-19}$ & $-0.971$ & $6.3{\times}10^{-7}$ \\
    \bottomrule
  \end{tabular*}
\end{table}

\begin{figure}[htbp]
  \centering
  \begin{minipage}[b]{0.48\columnwidth}
    \centering
    \includegraphics[width=\linewidth]{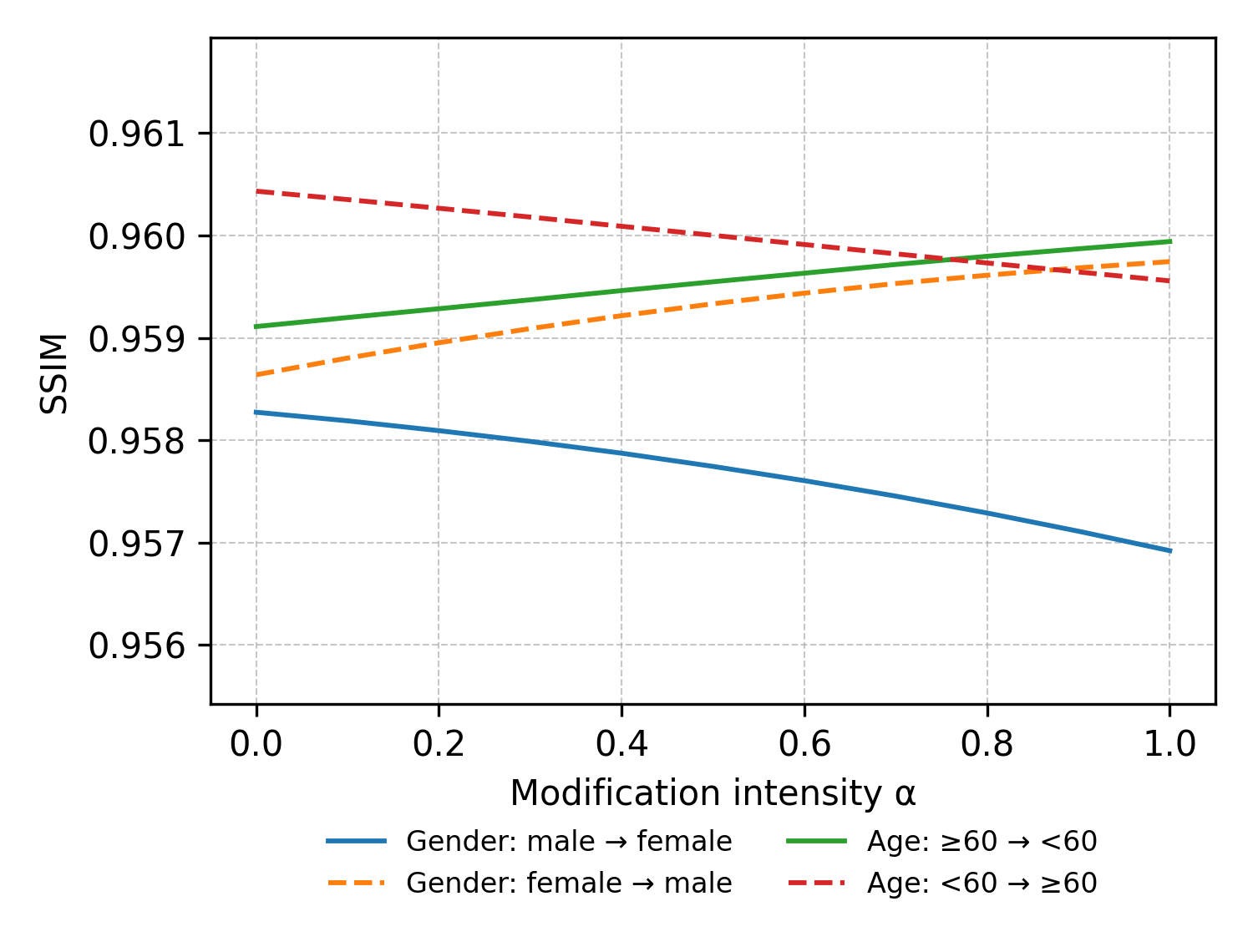}
    \caption*{(a) CNN encoder}
    \label{fig:ai_sex_curve}
  \end{minipage}
  \hfill
  \begin{minipage}[b]{0.48\columnwidth}
    \centering
    \includegraphics[width=\linewidth]{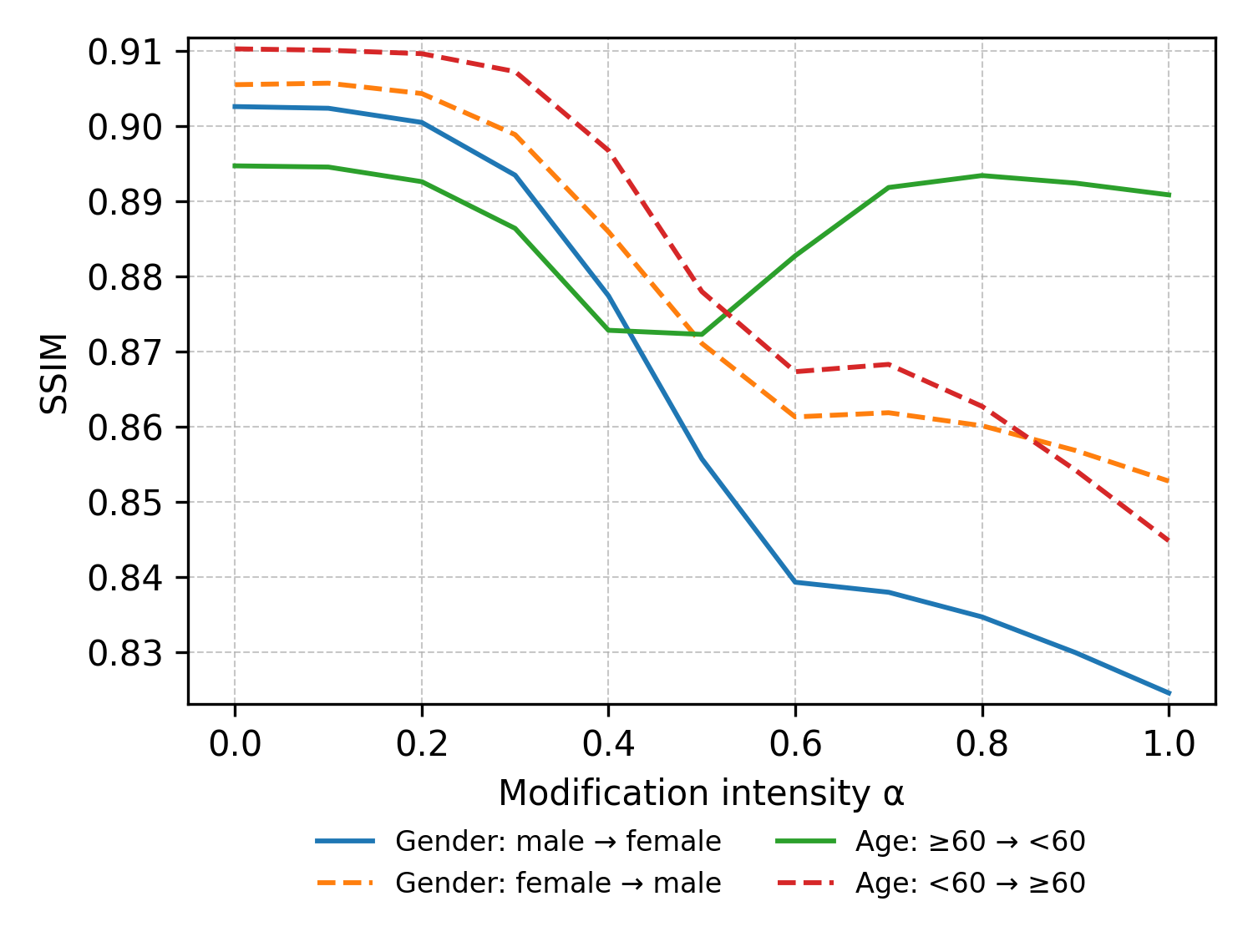}
    \caption*{(b) ViT encoder}
    \label{fig:ai_age_curve}
  \end{minipage}
  \caption{Bidirectional structural-similarity curves for the two
           attribute-neutralizers on ChestX-ray14.
           Each panel plots the mean SSIM between the original image and
           its edited counterpart for eleven edit intensities
           ($\alpha = 0.0$--$1.0$).
           Solid lines are edits applied to images that originally carry
           the attribute value 1 (male$\rightarrow$female or
           $\geq 60$\,y$\rightarrow$<$60$\,y). The ViT encoder produces markedly larger
           pixel-level changes at high $\alpha$ than the CNN baseline,
           especially in the male$\rightarrow$female and
           $\geq 60$\,y$\rightarrow$<$60$\,y directions.}
  \label{fig:ssim}
\end{figure}

\subsection{Attribute recognition by AI-Judge}

AI‐Judge performance was assessed on the test split for eleven neutralization intensities ($\alpha = 0.0, 0.1, \dots, 1.0$). Here, $\alpha = 0.0$ corresponds to the original image, while $\alpha > 0.0$ indicates outputs of the Attribute Neutralizer. Since the trained CNN neutralizer showed almost no change in leakage across all $\alpha$ levels, its leakage-metric curves and ROC curves have been moved to the Appendix. Here, we focus on the ViT neutralizer results, as they demonstrate a significant change in performance.

\subsubsection{Quantitative performance}

Leakage was assessed at three $\alpha$-levels (0.0, 0.5, 1.0) across three generator variants. The baseline U-Net reported by Hu et al.\ \cite{hu2024enhancing}, the ViT-adapted generator, and the trained CNN U-Net encoder. Results are shown in Table~2.

Table~2 shows that at the turning-point edit ($\alpha = 0.5$), the ViT neutralizer lowers the AI-Judge’s AUC by $\ge0.15$ relative to both Hu et al.’s U-Net and our own CNN retrain, while the CNN variant hardly changes. When pushed to $\alpha = 1.0$, AUC drops below 0.10, demographic cues are effectively flipped, and ACC, SEN, SPE, and F1 collapse in the same pattern. The re-trained CNN leaves all leakage metrics essentially intact.

\setlength{\lightrulewidth}{0.3pt}   
\setlength{\heavyrulewidth}{0.6pt}   

\begin{table}[htbp]
  \centering
  \setlength{\tabcolsep}{3pt}
  \small
  \caption{AI-Judge performance at $\alpha = 0.0$, $0.5$ and $1.0$. Rows are split by encoder.}
  \label{tab:ai_judge_performance}
  \begin{tabular}{lllrrrrr}
    \toprule
    Attribute & $\alpha$ & Encoder                & AUC    & ACC    & SEN    & SPE    & $F_1$  \\
    \midrule
    Sex       & 0.0      & U\mbox{-}Net$^{\star}$ & 1.00   & --     & --     & --     & --     \\
    Sex       & 0.0      & DeiT-S                & 0.9966 & 0.9684 & 0.9688 & 0.9680 & 0.9728 \\
    Sex       & 0.0      & U\mbox{-}Net          & 0.9966 & 0.9684 & 0.9690 & 0.9680 & 0.9730 \\
    \midrule
    Sex       & 0.5      & U\mbox{-}Net$^{\star}$ & 0.93   & --     & --     & --     & --     \\
    Sex       & 0.5      & DeiT-S                & 0.8315 & 0.7574 & 0.8176 & 0.6735 & 0.7969 \\
    Sex       & 0.5      & U\mbox{-}Net          & 0.9958 & 0.9661 & 0.9691 & 0.9619 & 0.9708 \\
    \midrule
    Sex       & 1.0      & U\mbox{-}Net$^{\star}$ & 0.08   & --     & --     & --     & --     \\
    Sex       & 1.0      & DeiT-S                & 0.0163 & 0.0692 & 0.0267 & 0.1285 & 0.0323 \\
    Sex       & 1.0      & U\mbox{-}Net          & 0.9956 & 0.9654 & 0.9690 & 0.9610 & 0.9700 \\
    \midrule
    Age       & 0.0      & U\mbox{-}Net$^{\star}$ & 0.91   & --     & --     & --     & --     \\
    Age       & 0.0      & DeiT-S                & 0.9413 & 0.8816 & 0.7592 & 0.9185 & 0.7480 \\
    Age       & 0.0      & U\mbox{-}Net          & 0.9413 & 0.8817 & 0.7590 & 0.9185 & 0.7480 \\
    \midrule
    Age       & 0.5      & U\mbox{-}Net$^{\star}$ & 0.84   & --     & --     & --     & --     \\
    Age       & 0.5      & DeiT-S                & 0.8008 & 0.6545 & 0.8356 & 0.5999 & 0.5283 \\
    Age       & 0.5      & U\mbox{-}Net          & 0.9357 & 0.8741 & 0.7543 & 0.9103 & 0.7351 \\
    \midrule
    Age       & 1.0      & U\mbox{-}Net$^{\star}$ & 0.17   & --     & --     & --     & --     \\
    Age       & 1.0      & DeiT-S                & 0.0710 & 0.1090 & 0.2689 & 0.0608 & 0.1226 \\
    Age       & 1.0      & U\mbox{-}Net          & 0.9326 & 0.8701 & 0.7430 & 0.9090 & 0.7260 \\
    \bottomrule
    \multicolumn{8}{p{0.93\linewidth}}{\footnotesize
      $^{\star}$Reported from Hu\,\textit{et al}. \cite{hu2024enhancing}}
  \end{tabular}
\end{table}

\subsubsection{Performance curves for patients’ age and sex attributes}\label{subsec:leakage_curve}

\begin{figure}[htbp]
  \centering
  \begin{minipage}[b]{0.48\columnwidth}
    \centering
    \includegraphics[width=\linewidth]{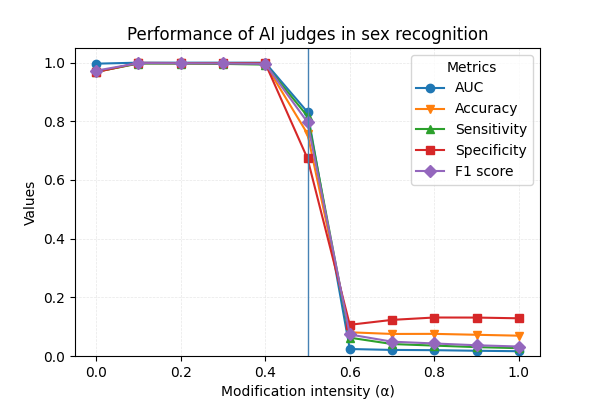}
    \caption*{(a) Attribute = sex}
  \end{minipage}
  \hfill
  \begin{minipage}[b]{0.48\columnwidth}
    \centering
    \includegraphics[width=\linewidth]{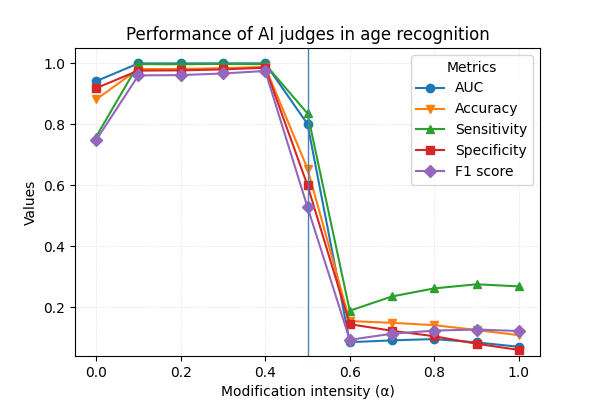}
    \caption*{(b) Attribute = age}
  \end{minipage}
  \caption{Leakage-metric curves for the ViT neutralizer. All metrics remain
           near 1.0 up to $\alpha = 0.4$ and then collapse sharply at
           $\alpha = 0.5$, matching the turning-point behaviour reported
           by Hu et al.\ \cite{hu2024enhancing}. Beyond $\alpha = 0.6$, the
           AI-Judge's AUC falls to $\leq 0.10$, showing that the images
           now encode the opposite demographic cues, while neutrality is
           achieved at $\alpha = 0.5$.}
  \label{fig:vit_leak_curves}
\end{figure}

Appendix Figures~\ref{fig:cnn_leak_sex}, \ref{fig:cnn_leak_age} (CNN encoder) and
Figure~\ref{fig:vit_leak_curves} (ViT encoder) plot the five leakage
metrics: AUC, Accuracy, Sensitivity, Specificity and $F_1$ against the
edit-intensity $\alpha$ for both attributes.
With the trained CNN neutralizer the curves remain almost flat, no turning point
is visible around $\alpha = 0.5$.
In contrast, the ViT curves reveal a steep downward trend, also noted by Hu et al.\ \cite{hu2024enhancing}. As $\alpha$ crosses 0.5, AUC drops to about 0.10, indicating the images now signal opposite sex and age cues.

\subsubsection{ROC overlays of AI-Judge for age and sex recognition}
\label{subsec:roc_overlays}

\begin{figure}[htbp]
  \centering

  \begin{minipage}[b]{0.48\linewidth}
    \centering
    \includegraphics[width=\linewidth]{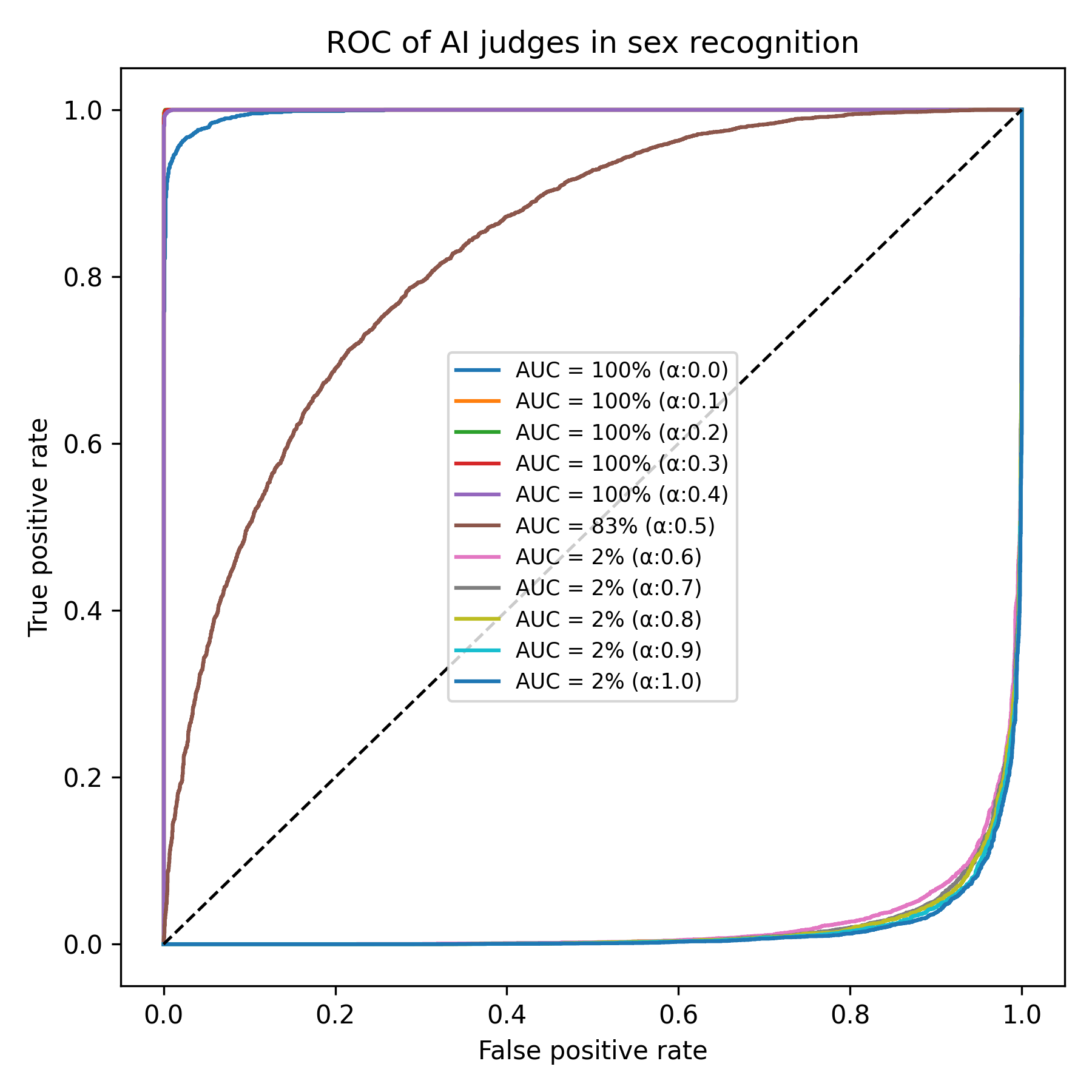}
    \caption*{(a) Protected attribute: sex (ViT)}
    \label{fig:vit_sex_roc}
  \end{minipage}
  \hfill
  \begin{minipage}[b]{0.48\linewidth}
    \centering
    \includegraphics[width=\linewidth]{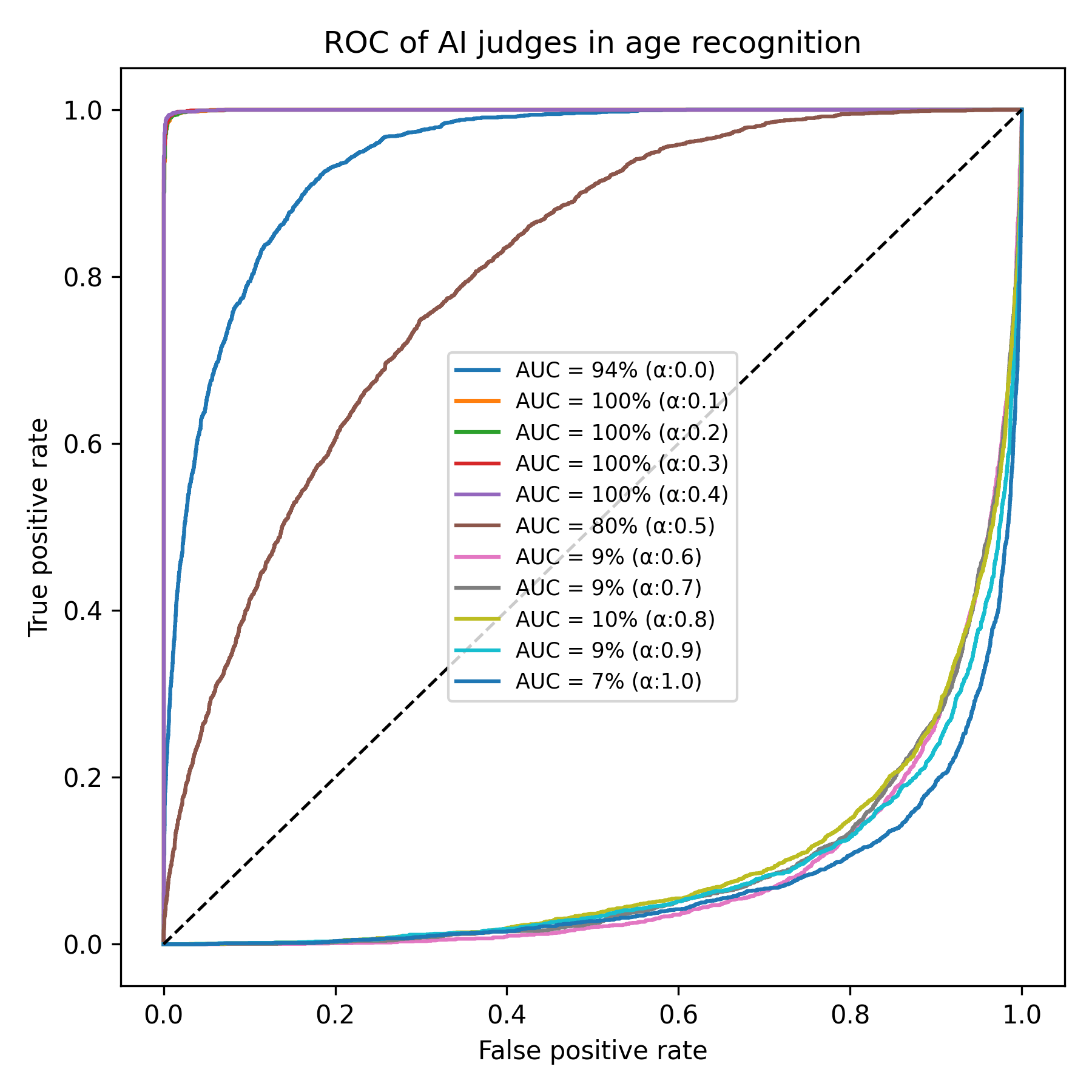}
    \caption*{(b) Protected attribute: age (ViT)}
    \label{fig:vit_age_roc}
  \end{minipage}

  \caption{Full-range ROC curves of \textbf{ViT-based} AI-Judges at the same
           eleven $\alpha$ levels. Neutralization has little effect up to
           $\alpha = 0.4$, but after the turning point ($\alpha \geq 0.5$) the
           curves rapidly approach the diagonal, signalling near-complete
           elimination of attribute leakage.}
  \label{fig:vit_roc_curves}
\end{figure}

Figures~\ref{fig:cnn_roc_sex}, \ref{fig:cnn_roc_age} (CNN encoder) and
Figure~\ref{fig:vit_roc_curves} (ViT encoder) compare the
ROC behaviour of the AI-Judge when applied
to \emph{original} images ($\alpha=0$) and to images neutralized with the eleven
modification intensities. For the CNN encoder, all curves remain
tightly clustered around the upper-left corner, resulting in
$\mathrm{AUC}\!\ge\!0.99$ for both sex and age recognition at every
$\alpha$. This shows that the CNN-based neutralizer fails to remove the
protected attributes.

In contrast, the ViT encoder behaves
identically to the original images up to $\alpha\!\le\!0.4$. At the turning
point $\alpha=0.5$ the ROC curves drop sharply (AUC  0.83 for sex and
0.80 for age) and for $\alpha\!\ge\!0.6$ they collapse towards the diagonal
(AUC $\le$ 0.10), indicating that attribute leakage is almost completely
suppressed. The residual spread is slightly larger for age than for sex,
reflecting the SSIM analysis.

\subsubsection{Grad-CAM heat activation map}

Figures~\ref{fig:gradcam_cnn} and \ref{fig:gradcam_vit} show the two
end-points (\(\alpha=0.0\) and \(0.5\)) for one representative
X-ray. The CNN‐based neutralizer leaves the hotspots largely
intact and increases the male probability from
\(p=0.29\) to \(p=0.75\).
In contrast, the ViT neutraliser erases the right-lung focus and
tightens the left-lung patch, pushing the classifier to an almost
certain male prediction (\(p=0.99\)). Figure~\ref{fig:gradcam_grid} presents the full $\alpha$-sweep for the ViT-based neutralizer, confirming that the highlighted sex evidence fades smoothly and the model's male probability falls toward 0.5 once $\alpha \ge 0.7$.

\subsection{Disease Diagnosis Model}

\subsubsection{Macro performance on ChestX-ray14}

Table~\ref{tab:singlecol_ddm} compares the macro diagnostic performance of the disease-diagnosis models trained on neutralized images from the CNN-based and ViT-based neutralizers. For models trained on original, un-neutralized images, the baseline macro ROC–AUC is 0.802. Neutralising with the CNN-based model results in a substantial loss in diagnostic utility, while the ViT-based neutralizer yields improvements across multiple metrics. Specifically, compared to the CNN baseline, the ViT neutralizer enhances macro ROC–AUC by approximately 0.28 for Sex and 0.26 for Age. 
The final AUC values for the ViT-based neutralizer (0.776 for Sex and 0.762 for Age) are only slightly below the un-neutralized baseline of 0.802, recovering 95–97\% of the diagnostic accuracy lost due to the neutralization process. The macro performance of other debiasing strategies is provided in Appendix~\ref{app:macro}.

\begin{table}[htbp]
\centering
\caption{Head-to-head macro diagnostic performance of the DeiT-Small neutralizer versus the CNN-based neutralizer on the common ChestX-ray14 test subset.}
\label{tab:singlecol_ddm}
\setlength{\tabcolsep}{5pt}
\small
\begin{tabular}{llccc}
\toprule
Attribute & Metric   & CNN   & ViT   & $\Delta$ (95\% CI)         \\
\midrule
Sex & ROC--AUC & 0.493 & 0.776 & +0.283 (0.271--0.296) \\
Sex & PR--AUC  & 0.084 & 0.203 & +0.119 (0.114--0.130) \\
Sex & ACC      & 0.235 & 0.678 & +0.444 (0.440--0.447) \\
Sex & SEN      & 0.831 & 0.726 & -0.105 (-0.122-- -0.089) \\
Sex & SPE      & 0.168 & 0.682 & +0.514 (0.511--0.518) \\
Sex & $F_1$    & 0.127 & 0.208 & +0.082 (0.077--0.086) \\
\midrule
Age & ROC--AUC & 0.503 & 0.762 & +0.259 (0.245--0.272) \\
Age & PR--AUC  & 0.086 & 0.196 & +0.110 (0.104--0.119) \\
Age & ACC      & 0.391 & 0.647 & +0.255 (0.251--0.260) \\
Age & SEN      & 0.600 & 0.709 & +0.109 (0.094--0.123) \\
Age & SPE      & 0.400 & 0.652 & +0.252 (0.248--0.256) \\
Age & $F_1$    & 0.068 & 0.200 & +0.132 (0.128--0.136) \\
\bottomrule
\end{tabular}
\normalsize
\vspace{2pt}
\footnotesize
$p$-values from 1\,000 bootstrap resamples are $<0.001$ for all metrics (two-sided).\\
\end{table}

\subsubsection{Per-finding level performance}

\begin{figure}[htbp]     
  \centering
  \includegraphics[width=0.8\linewidth]{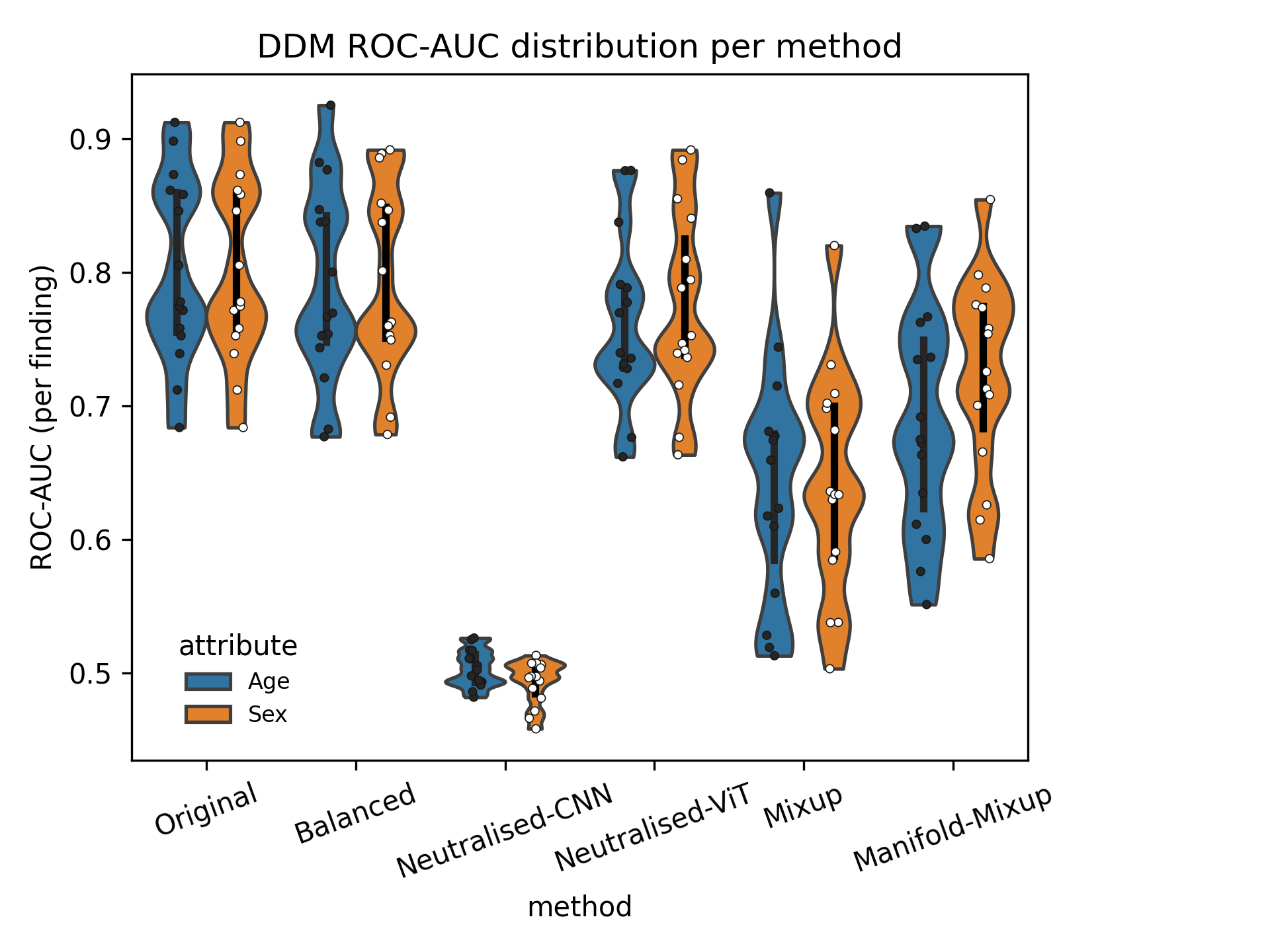}
  \caption{Per-finding ROC–AUC distributions (15 disease findings $\times$
           two attributes). White dots are individual findings; thick black
           bars denote the inter-quartile range, violin width is scaled
           equally across methods.}
  \label{fig:performance_perfinding}
\end{figure}

Figure~\ref{fig:performance_perfinding} illustrates the ROC–AUC distributions for the 15 disease findings across different debiasing methods. A global Friedman test indicates significant differences between methods ($p < 0.001$). The ViT-based neutralizer ranks higher than other methods, with its performance consistently surpassing the neutralized-CNN, which performs the worst.

The Nemenyi post-hoc test confirms that the neutralized–ViT neutralizer outperforms other methods, followed by Balanced Sampling and then Manifold-Mixup. The neutralized–CNN method ranks the lowest in performance, showing the least improvement in reducing bias across findings.
The corresponding average ranks and significant differences are summarised in the critical-difference diagram (Figure~\ref{fig:cd_diag}).

Pairwise DeLong comparisons for individual findings did not remain significant after FDR correction (all $p > 0.40$), likely due to the limited number of positive cases per finding. This highlights the class imbalance within the dataset, as identified in the exploratory data analysis for the ChestX-ray14 dataset (\ref{fig:cxr14_disease_freq}).

\begin{figure}[htbp]
  \centering
  \includegraphics[width=0.8\linewidth]{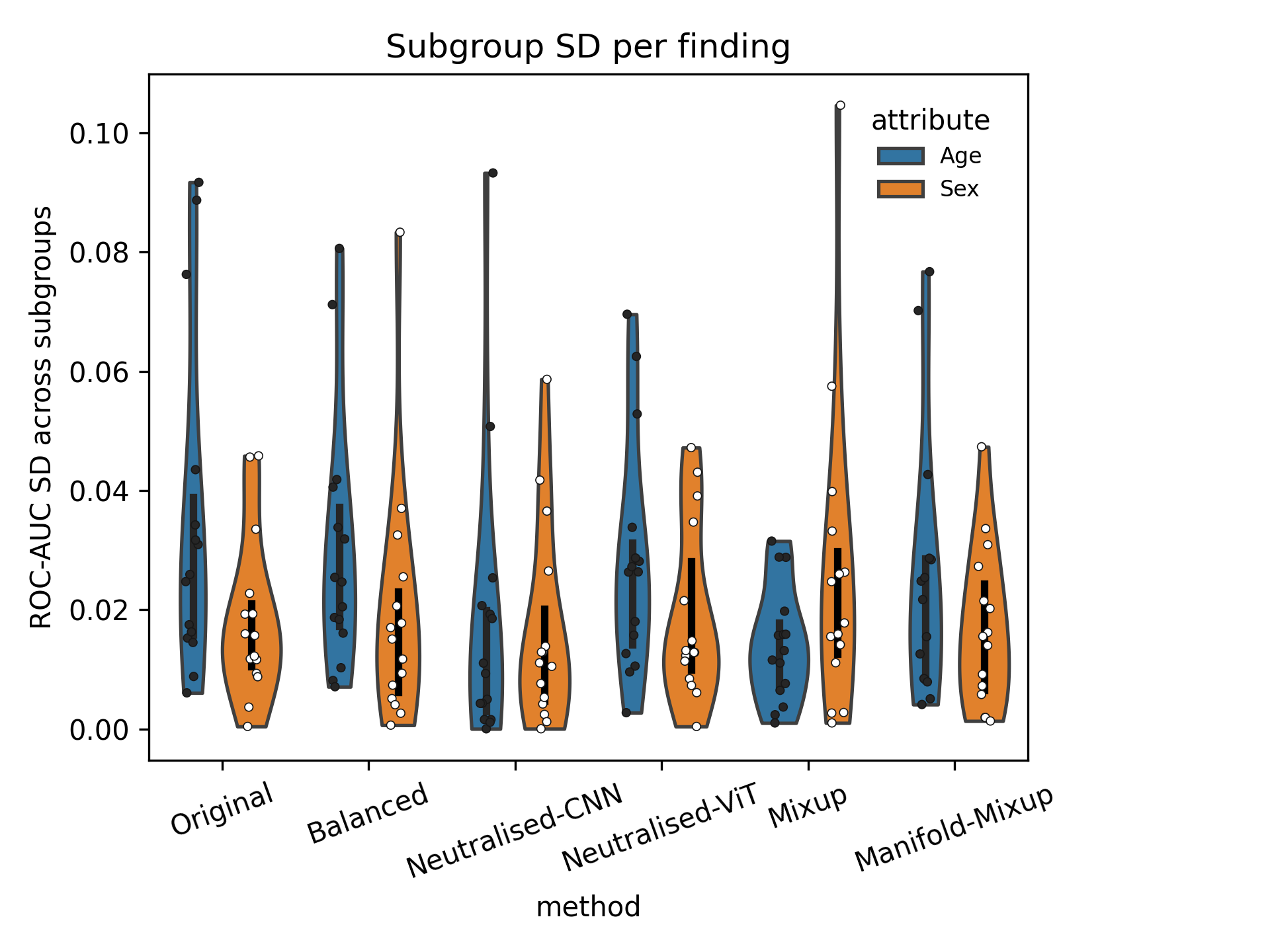}
  \caption{Sub-group standard deviation of ROC–AUC per finding.  
           White dots = individual findings; thick bars = inter-quartile range.}
  \label{fig:SD}
\end{figure}

\subsubsection{Unfairness across sub-groups}

Table~\ref{tab:unfairness_metrics} summarizes fairness metrics, including worst-case subgroup AUC, best–minus–worst AUC gap and subgroup-AUC standard deviation (SD) as medians over the 15 findings. Figure~\ref{fig:SD} illustrates the full distribution of SD values. A Friedman test finds no significant differences in SD across the six mitigation strategies ($\chi^2 = 5.87$, $p = 0.32$). However, a highly significant ranking for worst-case AUC ($\chi^2 = 133.1$, $p = 5.2\times10^{-27}$) places neutralized–CNN at the bottom. The ViT neutralizer performs similarly to the original and Balanced sampling methods for the best worst-case AUC, with the gap metric following the same pattern. For context, Table~\ref{tab:unfairness_metrics} also includes Hu et al.’s reported results\cite{hu2024enhancing}.

\begin{table}[ht]
  \centering
  \scriptsize
  \setlength{\tabcolsep}{2pt} 
  \caption{Sub-group unfairness on ChestX-ray14. Values are the median over 15 findings.}
  \begin{tabular}{l l r r r}
    \toprule
    Method                 & Attribute & Worst-case AUC & AUC Gap & AUC SD \\
    \midrule
    Original (baseline)    & Age       & 0.731          & 0.052   & 0.026  \\
                           & Sex       & 0.760 & 0.031   & 0.016  \\
    Balanced sampling      & Age       & 0.737 & 0.049   & 0.025  \\
                           & Sex       & 0.749          & 0.030   & 0.015  \\
    neutralized--ViT       & Age       & 0.698          & 0.053   & 0.026  \\
                           & Sex       & 0.746          & 0.026   & 0.013  \\
    Mixup                  & Age       & 0.609          & 0.026   & 0.013  \\
                           & Sex       & 0.628          & 0.036   & 0.018  \\
    Manifold-Mixup         & Age       & 0.638          & 0.043   & 0.022  \\
                           & Sex       & 0.711          & 0.031   & 0.016  \\
    neutralized--CNN       & Age       & 0.490          & 0.019   & 0.009  \\
                           & Sex       & 0.486          & 0.022   & 0.011  \\
    \midrule
    neutralized-CNN\,$^{\star}$ & Age  & 0.765          & 0.022   & 0.011  \\
    neutralized-CNN\,$^{\star}$ & Sex  & 0.776          & 0.006   & 0.003  \\
    \bottomrule
    \multicolumn{5}{p{\columnwidth}}{\footnotesize
      $^{\star}$Reported from Hu\,\textit{et al.} \cite{hu2024enhancing}.}
  \end{tabular}
  \label{tab:unfairness_metrics}
\end{table}

\subsubsection{Illustrative examples of attribute-neutralization}

Two qualitative views highlight how the neutralizers behave on individual images. Appendix Figures~\ref{fig:plate_cnn} and \ref{fig:plate_vit} show per-patient
AI-judge and diagnosis outputs before and after neutralization, contrasting
the CNN- and ViT-based pipelines. For each patient, the original CXR is displayed alongside its neutralized counterpart at \(\alpha=0.5\), together with AI-judge outputs (Age, Sex) and the five most likely findings predicted by the Disease Diagnosis Model.
Qualitatively, the CNN neutralizer suppresses attributes inconsistently and
distorts diagnosis scores, whereas the ViT neutralizer achieves stronger
attribute removal with minimal diagnostic drift.

\begin{figure}[htbp]
  \centering
  \includegraphics[width=0.9\linewidth]{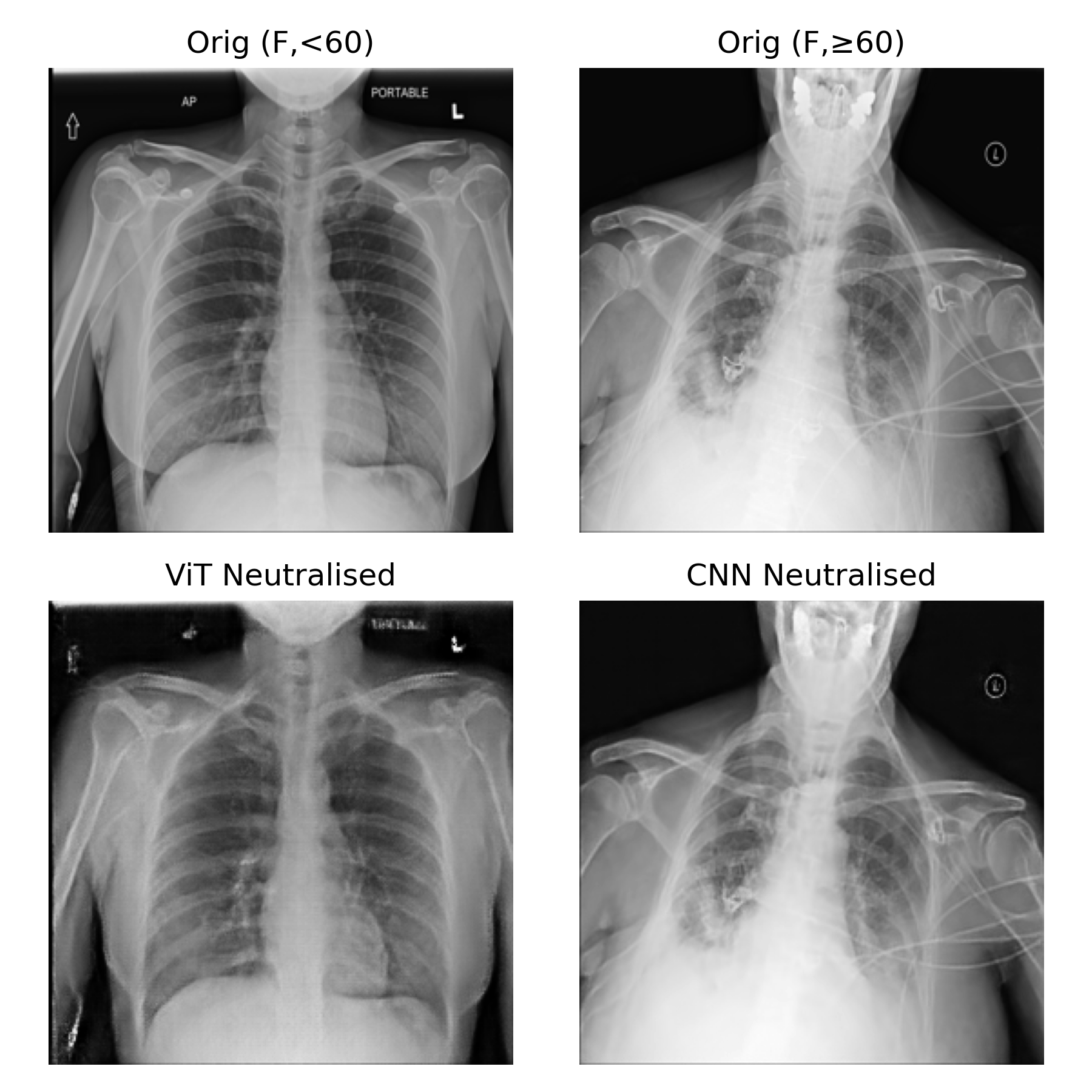}
  \caption{Side-by-side X-ray crops illustrating image-level edits.  
           Left: ViT neutralization visibly adjusts attribute-related anatomy while preserving overall appearance.  
           Right: CNN neutralization produces almost no visible change, consistent with its larger drop in DDM performance.}
  \label{fig:compare_cnn_vit}
\end{figure}

Figure~\ref{fig:compare_cnn_vit} zooms in on two example images (female \(<60\)\,y and female \(\ge60\)\,y) to visualise the edits themselves:  
ViT makes subtle changes around attribute-informative regions (breast tissue), whereas the CNN-based neutralizer leaves the image almost unchanged.

\section{Discussion}
\label{sec:discussion}

Replacing the GAN-based Attribute-Neutral Framework’s U-Net encoder with a DeiT-Small Vision Transformer cut sex- and age-leakage by about ten percentage points at the clinically relevant edit level (\(\alpha = 0.5\)) and, at stronger edits (\(\alpha \ge 0.6\)), pushed the AI-Judge AUC from approximately 0.99 down to \(\le 0.10\), all while preserving 95–97\% of baseline diagnostic ROC-AUC. These results demonstrate that attribute neutralisation generalises to self-attention backbones and maintains worst-case subgroup performance without clinically relevant loss of accuracy. This suggests that integrating global self-attention into pixel-level debiasing pipelines offers a practical path to fairer Chest X-ray AI systems.

\subsection{Comparison with previous literature}
The results extend Hu et al.'s\cite{hu2024enhancing}, who demonstrated fairness gains with CNN-based neutralizer but did not test non-CNN encoders. The turning-point behaviour reported (leakage collapses at $\alpha\!\approx\!0.5$) is confirmed, and a ViT backbone amplified this effect while largely retaining diagnostic utility. Prior fairness audits in radiology\cite{seyyed2021underdiagnosis,glocker2023risk} relied on post-hoc metrics or re-sampling, this work provides evidence that attribute neutralization remains effective when the encoder’s receptive field is global rather than convolution-local.

Strengths include the use of a held-out AI-Judge that quantifies sex- and age-leakage independently of diagnostic optimisation, a multi-modal evaluation design that cross-validates fairness via SSIM, AI-Judge metrics, subgroup AUC gaps, ROC overlays and Grad-CAM maps to provide converging evidence from complementary perspectives, and benchmarking against several alternative unfairness mitigation algorithms, which strengthens causal inference that the observed gains stem from the ViT neutralizer itself.

\subsection{Interpretation and mechanisms}
A plausible explanation for the sharp leakage drop in the ViT runs is the model’s global self-attention, which aggregates spatially dispersed attribute cues, such as breast tissue or rib-cage morphology into a single latent representation that can be overwritten when $\alpha\!\ge\!0.5$\cite{li2023transforming}. In contrast, U-Net skip connections are known to re-inject low-level textures\cite{he2019attgan}, if those textures encode sex or age information, the attribute vector may fail to cancel them fully. The SSIM analysis supports this: ViT edits are larger and more direction-symmetric than the CNN’s, indicating stronger control over attribute-bearing regions. However, the CNN baseline was trained for twice as many epochs and still followed Hu et al.’s fixed loss weights\cite{hu2024enhancing}. It therefore remains possible that additional tuning could narrow the ViT–CNN gap reported here, as the original Hu et al.\cite{hu2024enhancing} study achieved substantial leakage reduction with a U-Net CNN generator. Whether a transformer discriminator, diffusion-based generator\cite{weng2024fast}, or alternative GAN backbones such as StarGAN\cite{choi2018stargan} can further improve the fairness–utility curve remains an open question for future work.

\subsection{Clinical relevance}
Demographic bias in chest X-ray interpretation contributes to under-diagnosis in women, older adults and minority groups\cite{obermeyer2019dissecting}. In this study, the ViT neutralizer kept the worst-case AUC around 0.70 for every finding and slightly narrowed the subgroup AUC gap for sex (0.031 → 0.026) while leaving the gap for age unchanged (Table \ref{tab:unfairness_metrics}). This means the risk that any subgroup receives systematically poorer care is reduced, or at least not increased relative to the baseline model. Because the framework edits images directly, the neutralized X-rays remain readable to radiologists and can be stored alongside the originals for audit. Showing the two versions side-by-side in reporting workstations would make the fairness intervention transparent and open to clinical scrutiny.

Designing a socially acceptable system also requires stakeholder input\cite{liu2023translational,ricci2022addressing}. Clinicians and patients could help decide how much attribute removal is tolerable before image quality or personal identity feels compromised, guiding site-specific choices of the edit-intensity parameter~$\alpha$.

\section{Conclusion}
\label{sec:conclusion}

This work set out to determine the impact of replacing the U-Net encoder in Hu et al.’s Attribute-Neutral Framework\cite{hu2024enhancing} with a Vision-Transformer encoder when diagnosing diseases on chest-X-rays. the ViT extension significantly reduces sex- and age-related attribute leakage while retaining 95–97\,\% of baseline diagnostic ROC-AUC, thereby demonstrating a favourable fairness–performance trade-off. Relative to the original model, worst-case subgroup performance is preserved and the sex-based AUC gap is modestly reduced.

The findings confirm that pixel-level attribute neutralization can improve fairness without a clinically relevant loss of accuracy, at least on the ChestX-ray14 dataset. Because the edits remain visible to clinicians and can be displayed beside the original image, the pipeline lends itself to participatory evaluation, side-by-side review and continuous leakage monitoring—practical features that can foster social acceptability beyond metric parity (Sub-question 3).

Four limitations qualify these conclusions. First, all experiments were confined to one public dataset; external validity across hospitals, modalities and attributes such as race or body-mass index is untested. Second, the ViT was trained under a reduced compute budget and inherited fixed loss weights from the CNN baseline, adaptive weighting or longer training may provide further gains. Third, only imaging-level fairness was considered, downstream clinical decision impact was beyond scope.
Finally, all runs for the ViT used a single random seed (0) due to GPU quota limits, which may impact the reproducibility and variance of the results. Future work should average across multiple random seeds to ensure greater reliability and report variance.

Within these bounds, this study advances the state of the art by demonstrating that self-attention backbones can underpin pixel-space debiasing and that attribute neutralization remains effective when the encoder’s receptive field is global rather than convolution-local.  The results encourage broader exploration of transformer-based neutralizers and discriminators, adaptive loss balancing, and multi-site validation. The next step for future researchers is to repeat the experiment on data from diverse hospitals and scanners, while re-tuning the edit-intensity parameter $\alpha$ and the loss-weight schedule. This would clarify whether ViT-based neutralization can generalise beyond ChestX-ray14 and remain practical in routine AI healthcare workflows.

This research contributes to the advancement of fair AI in healthcare by demonstrating that the integration of self-attention-based architectures, specifically through the hybrid ViT encoder in the Attribute-Neutral Framework, effectively mitigates sex- and age-related bias in chest X-ray diagnostics. The findings show that the ViT-based neutraliser improves attribute recognition performance while maintaining 95-97\% of baseline diagnostic accuracy, despite requiring half of the training epochs. These results offer a promising pathway for the development of more robust, fair, and clinically applicable AI systems in medical imaging, while also highlighting key limitations that future research can address to further enhance fairness and utility in diverse clinical environments.

\section*{Declarations}

\noindent\textbf{Data Availability:}
All experiments use publicly available, de-identified chest X-ray datasets. The primary dataset is ChestX-ray14 from the U.S. National Institutes of Health (NIH). Additional exploratory analyses use CheXpert, MIMIC-CXR, and PadChest. Access to these datasets is governed by their respective data-use agreements. Pre-processing scripts and model training code (and, where possible, trained model weights) are available from the corresponding author upon reasonable request.

\medskip
\noindent\textbf{Funding:}
No external funding was received for this study.

\medskip
\noindent\textbf{Ethics and Consent to Participate:}
Not applicable. The work relies exclusively on publicly available, fully de-identified chest X-ray datasets and does not involve any new data collection, intervention, or contact with human participants or animals.

\medskip
\noindent\textbf{Consent to Publish:}
Not applicable.

\medskip
\noindent\textbf{Competing Interests:}
The authors declare that they have no competing interests.

\newpage
\onecolumn

\bibliography{references.bib}

\clearpage
\appendix
\onecolumn

\section{Dataset Pre-processing}
\label{app:preproc}

\subsection{ChestX-ray14}\label{app:pre_cxr14}

The NIH metadata flags frontal views so no additional filtering is needed. The fourteen findings are one-hot encoded and any blank or uncertain annotations are mapped to zero. After demographic cleaning the cohort includes 112 120 images from 30 805 patients.

\subsection{MIMIC-CXR}\label{app:pre_mimic}
Filtering for AP/PA views reduces the total of 377\,110 images to 243\,334. The dataset includes 13 findings, and removing missing or $-1$ labels results in 202\,306 images from 47\,699 patients. Each finding can have one of four possible labels: ``positive'', ``negative'', ``not mentioned'', or ``uncertain''. The last three options are grouped under the negative label. The ``Unknown'' category for race is removed, with patients being classified as ``White'', ``Hispanic'', ``Black'', ``Asian'', ``American Indian'', or ``Other''. The ``Unknown'' category for insurance type is removed, grouping patients as ``Medicaid'', ``Medicare'', or ``Other''.

\subsection{CheXpert}\label{app:pre_chexpert}
Lateral views are removed, leaving 191 026 frontal images from 64 533 patients. Each finding can have one of four labels: “positive,” “negative,” “not mentioned,” or “uncertain.” Likewise, the last three options are grouped under the negative label.

\subsection{PadChest}\label{app:pre_padchest}
AP and PA projections are selected and patient age is calculated from birth year, resulting in 96 274 images from 63 938 patients.

\section{Exploratory Data Analysis}
\label{app:eda}

\subsection{ChestX-ray14}\label{app:eda_cxr14}

Age-balanced mini-batch sampling.
A \mbox{$\chi^{2}$} test indicates a significant imbalance (\mbox{$\chi^{2}$} = 31 304.4, p < 0.001), with patients aged $\geq 60$ accounting for only 23.58\% of X-ray images (\ref{fig:cxr14_demographics}). 

\begin{figure}[p]
  \centering
  \includegraphics[width=\textwidth]{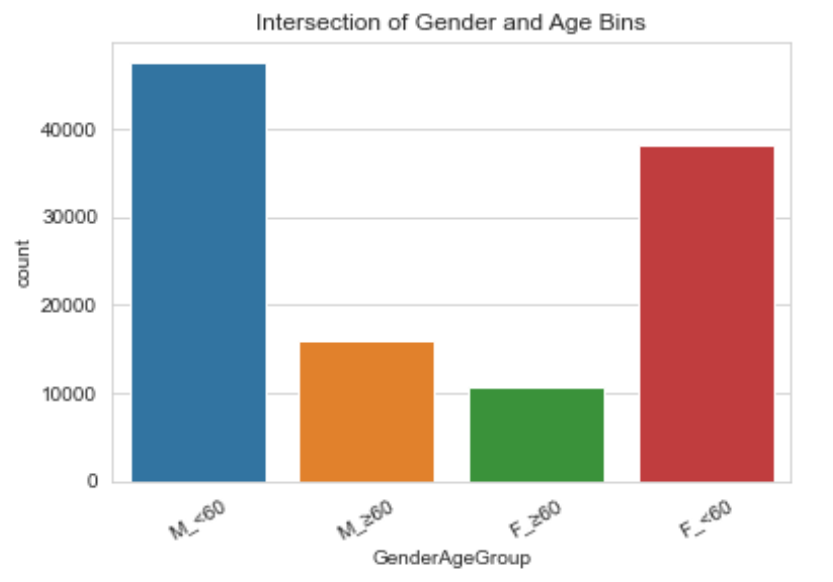}
  \caption{
  Demographic composition of ChestX-ray14.
  Left: patient‐sex counts; right: counts after binning age into
  “$<$60’’ and “$\ge 60$’’.
  A $\chi^{2}$ goodness-of-fit test rejects the hypothesis of equal
  class frequencies ($p<0.001$), confirming substantial imbalance in
  both protected attributes.
}
  \label{fig:cxr14_intersection}
\end{figure}

Up‐weighting of subgroup older women.
A Wilson score confidence interval for the proportion of female patients aged $\ge60$ is 9.41\% (95\% CI: 9.24–9.58\%), showing that this subgroup is the smallest (\ref{fig:cxr14_intersection}).

Long‐tail aware loss.
Label frequencies range from {No Finding} at 60 361 images (42.7\%) to \textit{Hernia} at 227 images (0.16\%), spanning nearly three orders of magnitude (\ref{fig:cxr14_disease_freq}).

\clearpage

\clearpage
\subsection{MIMIC-CXR}\label{app:eda_mimic}

\subsubsection{\textbf{MIMIC-CXR}}

\begin{figure}[p]
  \centering
  \includegraphics[width=\textwidth]{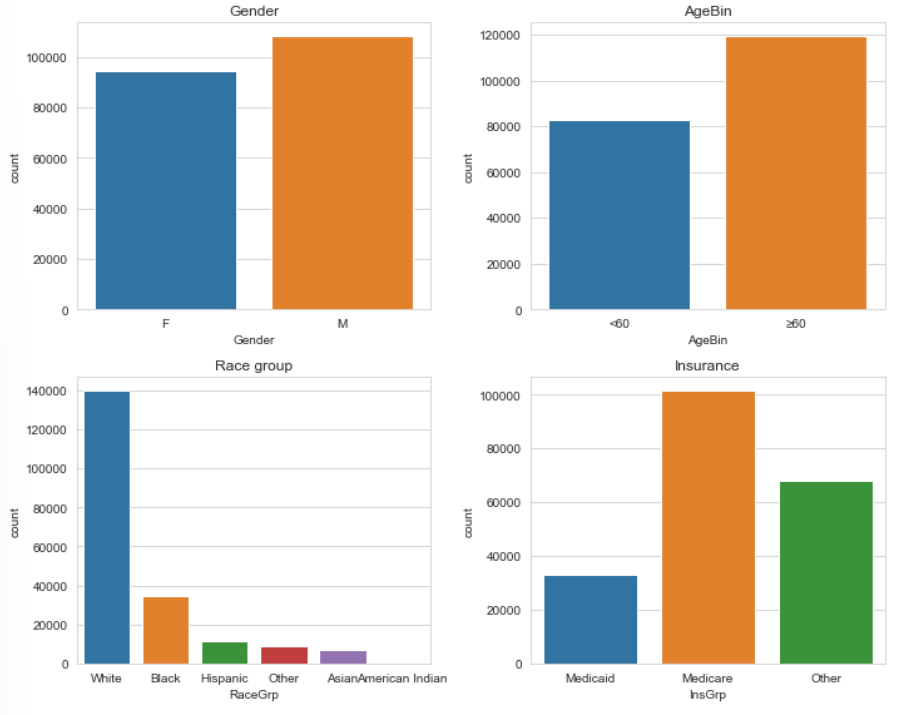}
  \caption{%
  Demographic breakdown of the MIMIC-CXR training set.
  Top: distributions of patient sex and of age binned at 60 years.
  Bottom: distributions of self-reported race and of admission‐insurance
  group.
  A $\chi^{2}$ test confirms significant imbalance in each attribute,
  most notably a majority of patients aged $\ge 60$ and a strong
  skew towards the “White’’ race group.%
}
  \label{fig:mimic_demographics}
\end{figure}

\begin{figure}[p]
  \centering
  \includegraphics[width=\textwidth]{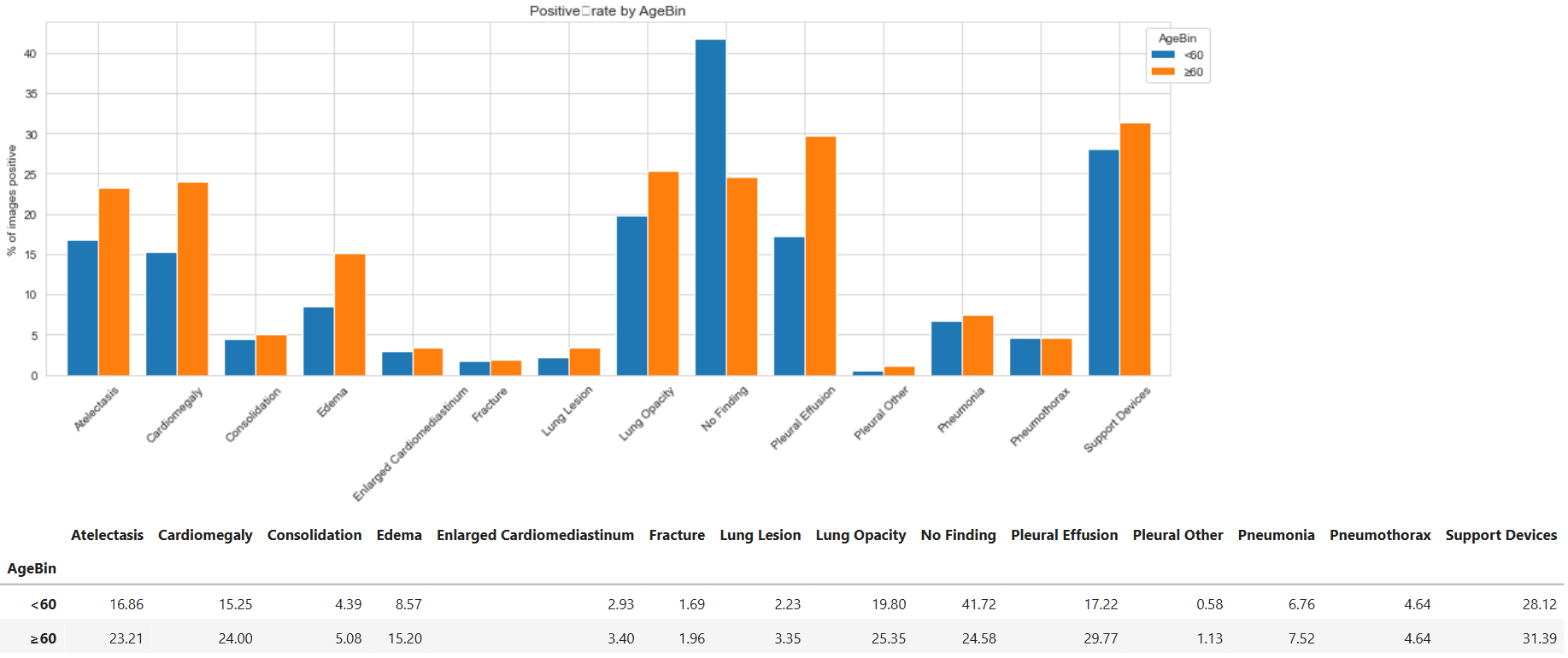}
  \caption{
  Per-finding prevalence stratified by age.
  Several cardiopulmonary conditions (e.g.\ Pleural Effusion,
  Cardiomegaly) are markedly more common among patients aged
  $\ge 60$, whereas the “No Finding’’ label is less common,
  illustrating the covariate shift that motivates age-aware sampling.
}
  \label{fig:mimic_posrate}
\end{figure}

Age-aware sampling.
Older patients dominate within the MIMIC-CXR dataset with 59\% of the images originating from patients aged $\ge60$ against 41\% $<60$. A \mbox{$\chi^{2}$} test confirms the imbalance ($\chi^{2}=4\,665$, $p<0.001$, $V=0.184$;)( \ref{fig:mimic_demographics}). 

image-level positive-rate-bar-plot.
Label prevalence also shifts substantially with age. \textit{Pleural Effusion} rises from 17.2\% in younger patients to 29.8\% in seniors, \textit{Cardiomegaly} from 15.3\% to 24.0\%, while \textit{No Finding} drops by 17 pp (\ref{fig:mimic_posrate}).

\begin{figure}[p]
  \centering
  \includegraphics[width=\textwidth]{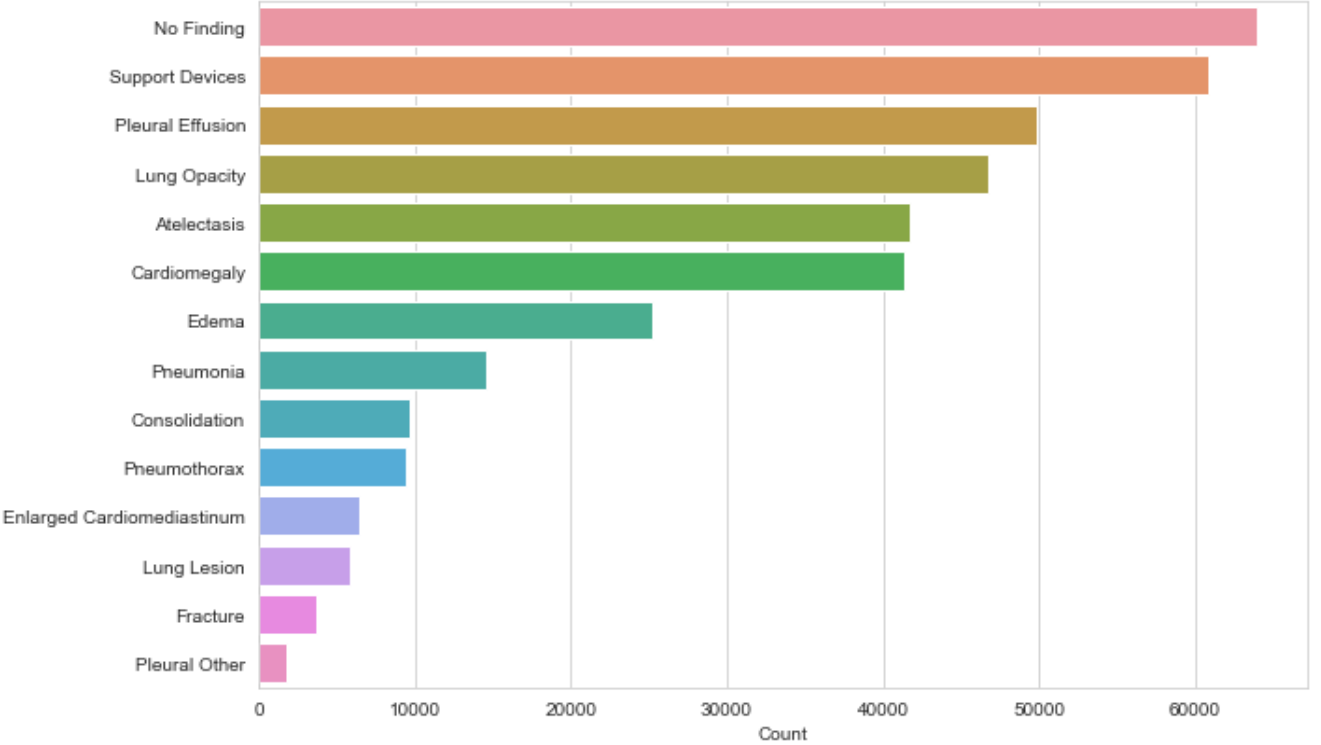}
  \caption{
  Overall label frequencies in MIMIC-CXR.
  Two high-level categories (“No Finding’’ and “Support Devices’’)
  dominate, while multiple pathologies occur only rarely,
  yielding a head–tail ratio of more than one order of magnitude
  that necessitates long-tail-aware loss weighting.
}
  \label{fig:mimic_disease_counts}
\end{figure}

Head–tail label imbalance.
\textit{No Finding} (31.6\%) and \textit{Support Devices} (30.1\%) dominate rare labels such as \textit{Pleural Other} (0.9\%) and \textit{Fracture} (1.8\%), resulting in a 35× prevalence spread (\ref{fig:mimic_disease_counts}). 

\clearpage

\clearpage

\clearpage

\clearpage
\subsection{CheXpert}\label{app:eda_chexpert}

\begin{figure}[p]
  \centering
  \includegraphics[width=\textwidth]{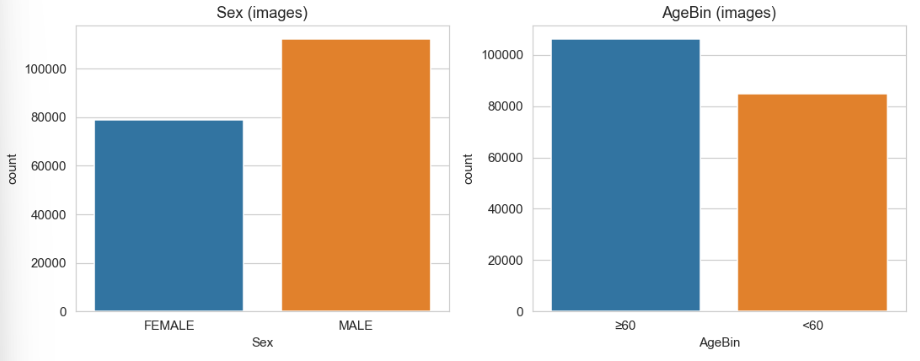}
  \caption{
Demographic composition of the CheXpert training set.
Left: sex distribution of X-ray images; right: age split at 60 years.
A $\chi^{2}$ test confirms a pronounced imbalance, with younger
patients substantially out-numbering seniors—hence the use of
age-balanced mini-batch sampling in our experiments.
}
  \label{fig:chexpert_demographics}
\end{figure}

Age-balanced mini-batch sampling is motivated by a strong class imbalance in the age bins. An age‐bin \mbox{$\chi^{2}$} test shows that patients aged $\ge 60$ represent only 33\,\% of images, while those $<60$ account for 67\,\%. The deviation from a 50–50 split is highly significant ($\chi^{2}\!\approx\!2.1\times10^{4}$, $p<0.001$; Figure~\ref{fig:chexpert_demographics}).

\begin{figure}[p]
  \centering
  \includegraphics[width=\textwidth]{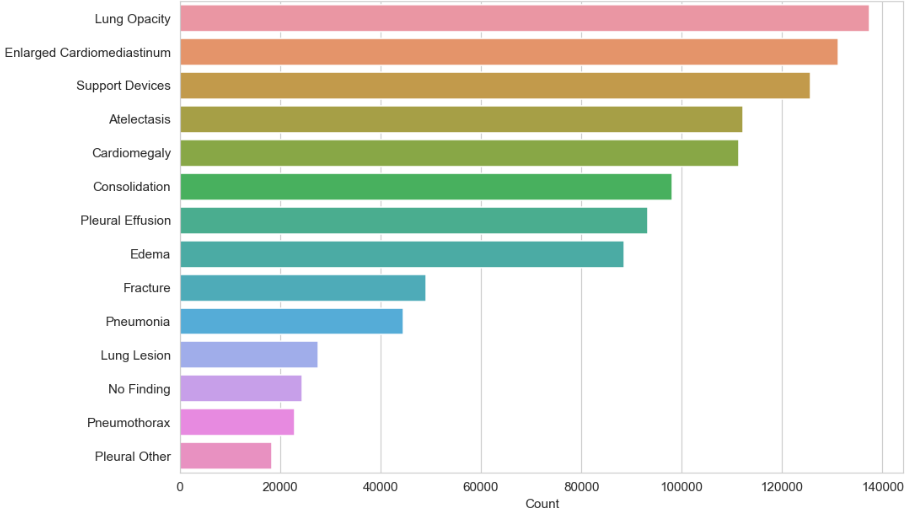}
  \caption{
Image-level frequencies of the 14 diagnostic labels in CheXpert.
The class hierarchy is long-tailed: common findings such as
Lung Opacity and Enlarged Cardiomediastinum dominate, whereas rare
categories (e.g.\ Pleural Other) occur an order of magnitude less
often.
This motivates the long-tail-aware loss weighting applied during
classifier training.
}
  \label{fig:chexpert_disease_counts}
\end{figure}

A long-tail aware loss is required because the dataset's label frequencies are highly imbalanced. For example, \textit{Lung Opacity} appears in 137\,342 images (72 \%), \textit{Pneumothorax} in 22\,804 (12 \%), and \textit{Pleural Other} in 18\,387 (9.6 \%), giving a head–tail prevalence spread of approximately $\sim$7.5$\times$ (Figure~\ref{fig:chexpert_disease_counts}).

\clearpage

\clearpage

\clearpage
\subsection{PadChest}\label{app:eda_padchest}

PadChest is the only large European CXR dataset in the study. Its images originate from a different health-care system in Spain, making the dataset a way to evaluate whether fairness interventions generalise beyond U.S. patient populations.

\begin{figure}[p]
  \centering
  \includegraphics[width=\textwidth]{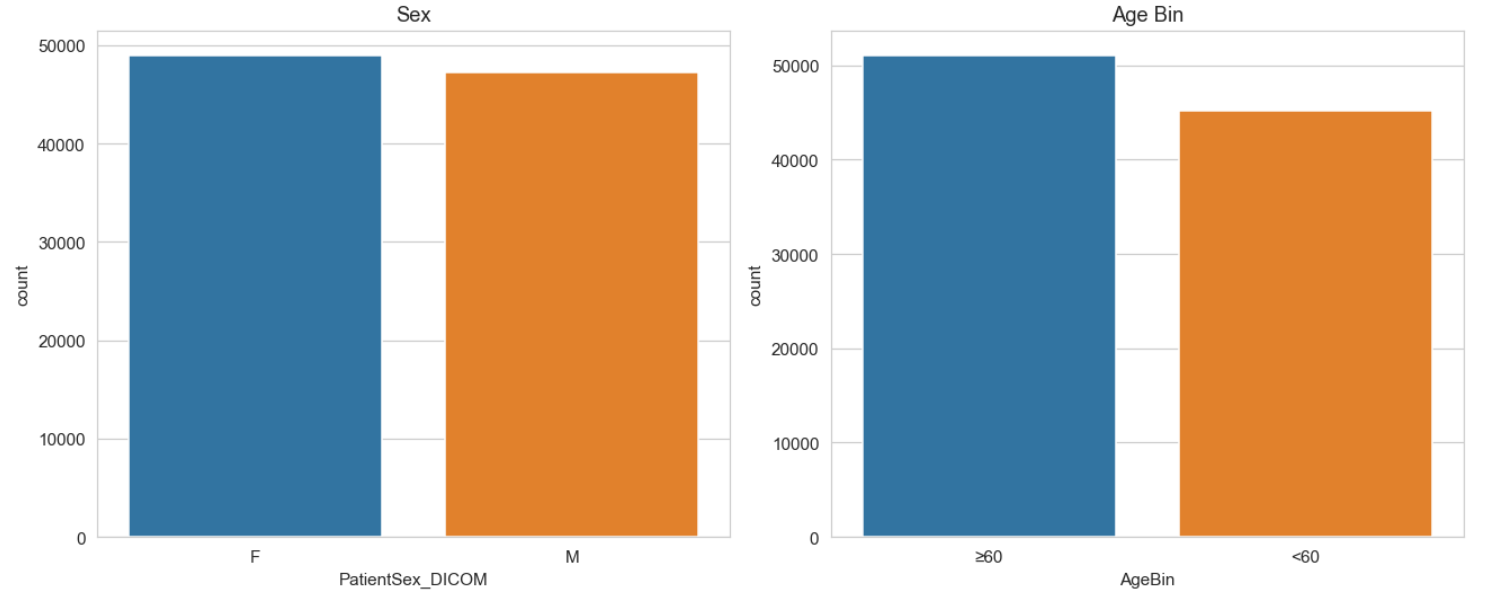}
  \caption{
Sex (left) and age-bin (right) counts for PadChest.
The dataset is slightly male-skewed and contains a majority of
patients aged $\ge 60$.
A $\chi^{2}$ test confirms significant age imbalance, motivating
age-aware sampling in later experiments.
}
  \label{fig:padchest_demographics}
\end{figure}

Age-dominated prevalence shifts.
Older patients make up 53\% of the dataset (Appendix Figure 10). The \mbox{$\chi^{2}$} test results in $\chi^{2}=21\,074$, $p<0.001$, Cramér’s $V=0.536$, indicating that the attribute age explains more than 25\% of label variance. At the image level (Appendix Figure 11), the \textit{Normal} label falls from 54.3\% in younger patients to 20.2\% in seniors, \textit{Cardiomegaly} rises from 2.1\% to 14.7\%, and \textit{COPD} increases from 6.8\% to 20.7\% (\ref{fig:padchest_demographics}).

\begin{figure}[p]
  \centering
  \includegraphics[width=\textwidth]{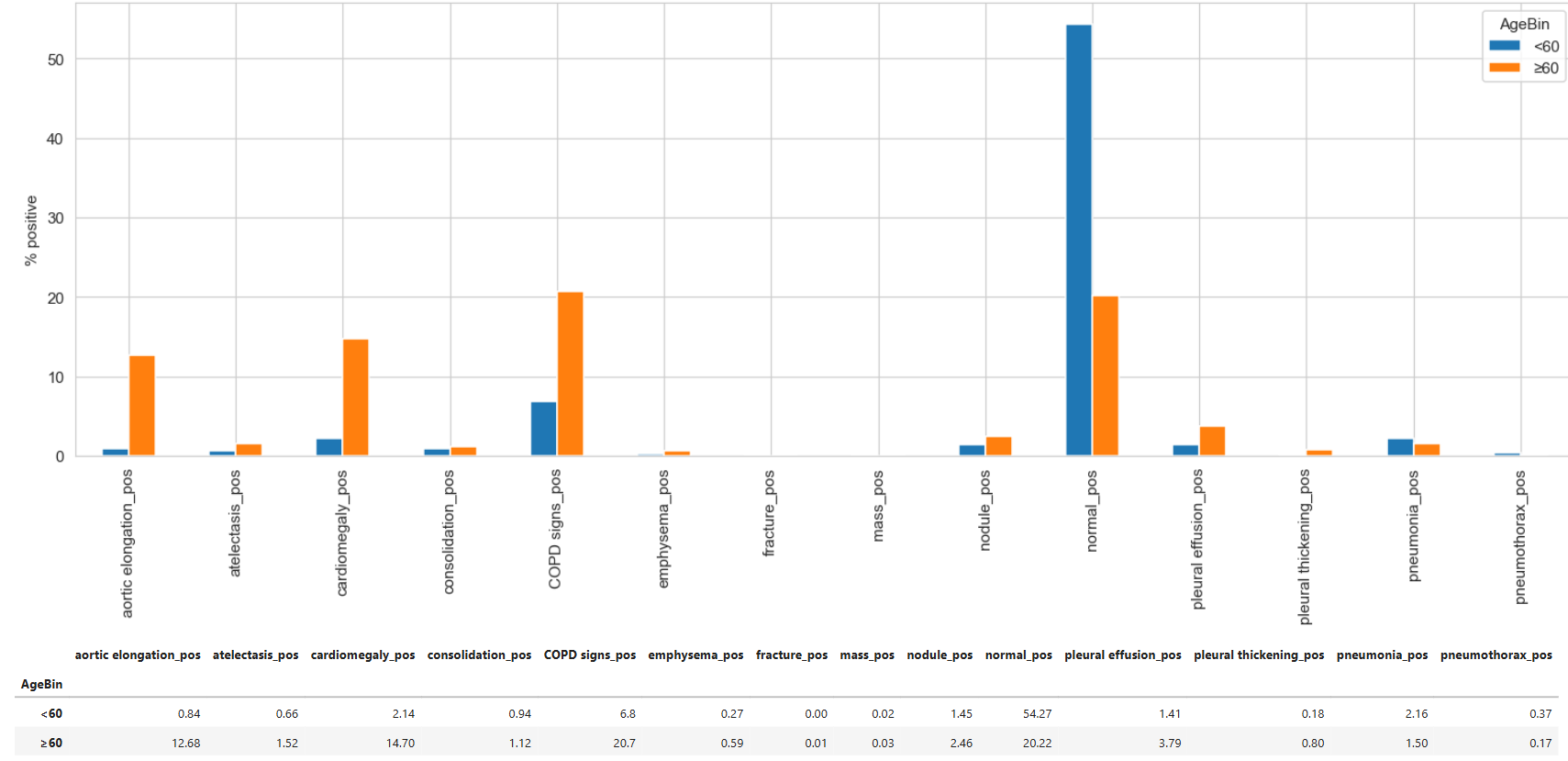}
  \caption{
Prevalence of 16 diagnostic labels stratified by age.
Several findings (e.g.\ Cardiomegaly, COPD signs) are markedly more
common in seniors, whereas the “Normal’’ label is far more prevalent
in younger patients, illustrating an age-driven covariate shift.
}
  \label{fig:padchest_posrate1}
\end{figure}

\begin{figure}[p]
  \centering
  \includegraphics[width=\textwidth]{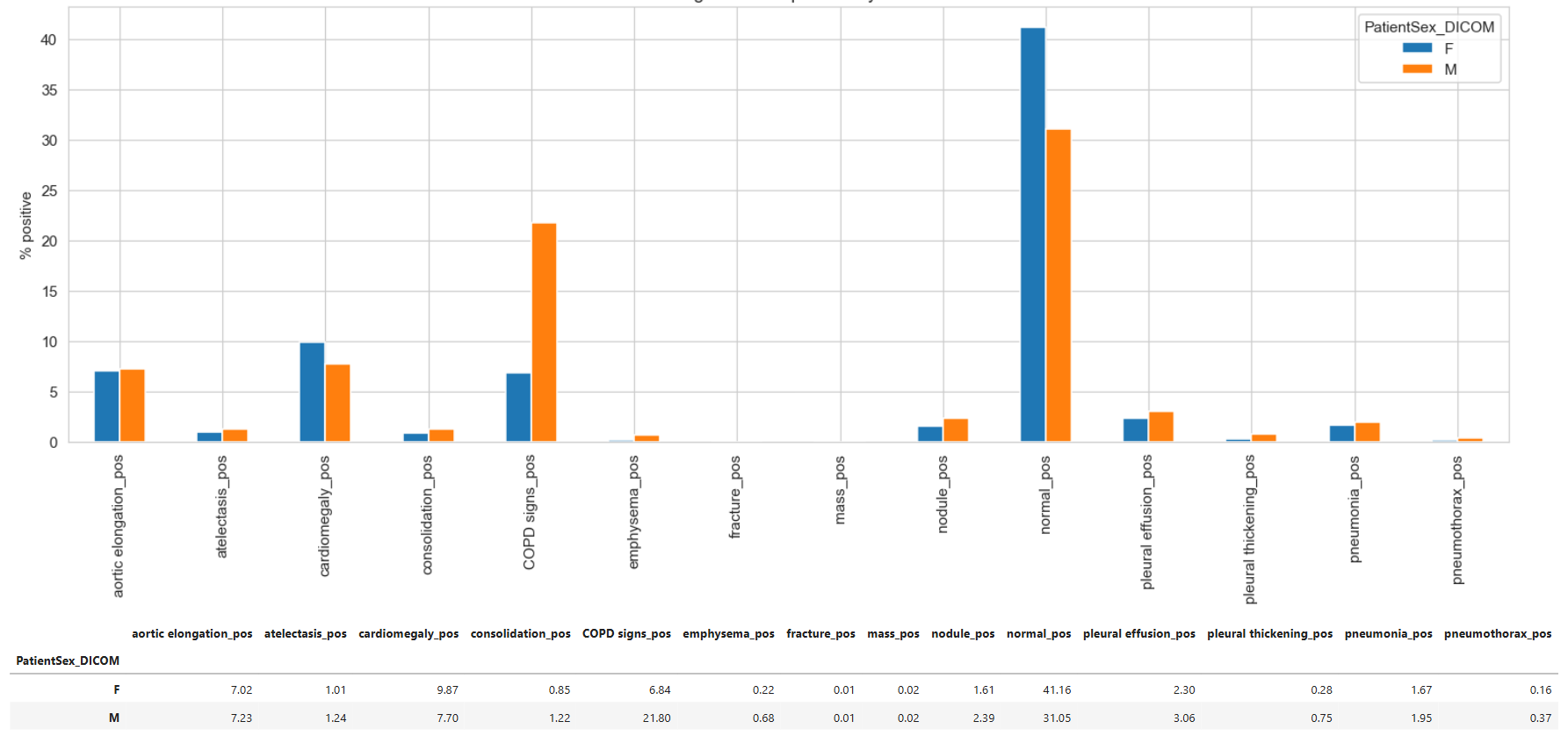}
  \caption{
Same prevalence plot, now stratified by sex.
COPD signs appear more than three times as often in men, while the
“Normal’’ label is considerably higher in women.
These divergences underpin the potential for sex bias in naïve
classifiers.
}
  \label{fig:padchest_posrate2}
\end{figure}

Male-skewed COPD and Normal labels
\textit{COPD} signs appear in 21.8 \% of male against 6.8 \% of female images. Conversely, the label \textit{Normal} is 41.2 \% in females and 31.0 \% in males. A contingency test over all labels and sex results in $\chi^{2}=4\,915$ ($p<0.001$, $V=0.259$;)(\ref{fig:padchest_posrate1}, \ref{fig:padchest_posrate2}). 

\begin{figure}[p]
  \centering
  \includegraphics[width=\textwidth]{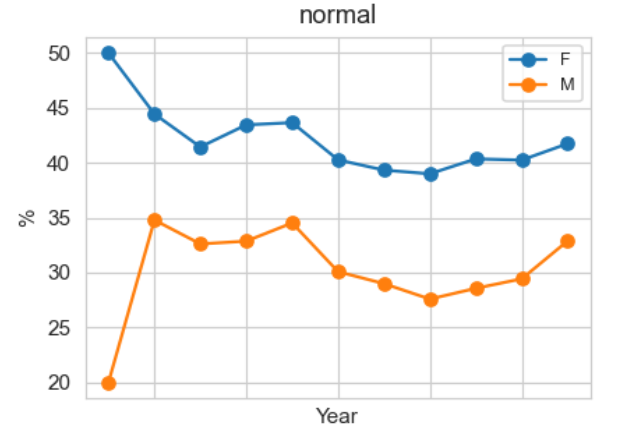}
  \caption{
Year-over-year proportion of images labelled “Normal”, separated by
sex. Throughout the 2009–2017 period women consistently show a
higher normal rate than men by roughly 15–20 percentage points,
highlighting a persistent sex gap.
}
  \label{fig:padchest_trend_normal}
\end{figure}

\begin{figure}[p]
  \centering
  \includegraphics[width=\textwidth]{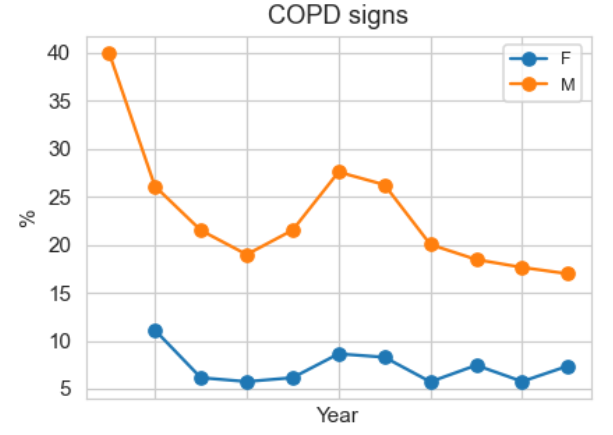}
  \caption{
Temporal evolution of the prevalence of “COPD signs’’ in PadChest,
split by sex. Men exhibit a substantially higher and more variable
rate than women across all years, reinforcing the male-skew observed
in the cross-sectional analysis.
}
  \label{fig:padchest_trend_copd}
\end{figure}

Temporal trends further highlight a gender gap between the \textit{Normal} and \textit{COPD} labels. Between 2009 and 2017, women show the \textit{Normal} label roughly 20 percentage points more often than men, while men exhibit \textit{COPD} labels about 16 percentage points more frequently than women (Figures~\ref{fig:padchest_trend_normal} and \ref{fig:padchest_trend_copd}). Without mitigation, a model trained on this data could learn spurious associations such as “male implies COPD” and “female implies Normal.”

\clearpage

\clearpage

\clearpage

\clearpage

\clearpage

\clearpage
\section{Supplementary Model Figures}
\label{app:modelfigs}

\clearpage
\subsection{Leakage-metric curves (CNN)}\label{app:leak_curves}
\begin{figure}[htbp]
  \centering
  \includegraphics[width=\textwidth]{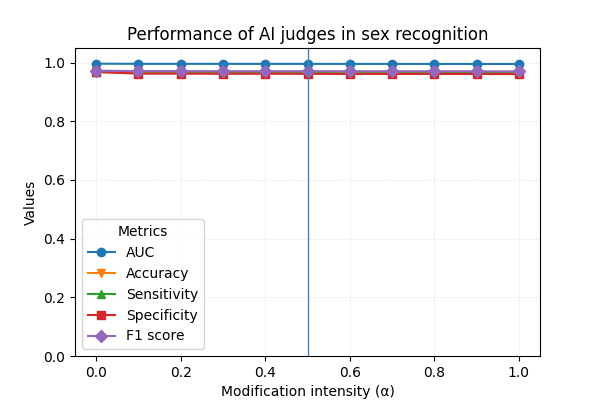}
    \caption{
  Performance of the sex-leakage judge as a function of the CNN
  neutralisation weight $\alpha$.
  All five metrics stay flat at their
  $\alpha = 0$ value (AUC and ACC $\approx 1.00$), indicating that the CNN edit
  fails to suppress sex information at any strength. The vertical line at
  $\alpha = 0.5$ marks the operating point discussed in the main text.
}

  \label{fig:cnn_leak_sex}
\end{figure}

\clearpage
\begin{figure}[htbp]
  \centering
  \includegraphics[width=\textwidth]{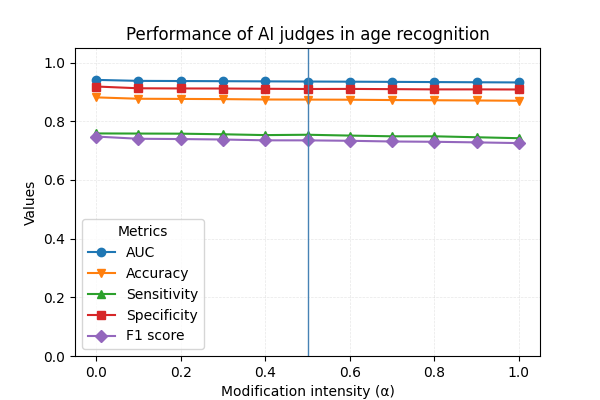}
  \caption{
  Performance of the age-leakage judge versus neutralisation strength for
  the same CNN encoder.  Metrics decrease only marginally even at
  \(\alpha=1.0\), showing that age cues are also largely retained.
}
  \label{fig:cnn_leak_age}
\end{figure}

\clearpage
\subsection{ROC overlays (CNN)}\label{app:roc_overlays}
\begin{figure}[htbp]
  \centering
  \includegraphics[width=\textwidth]{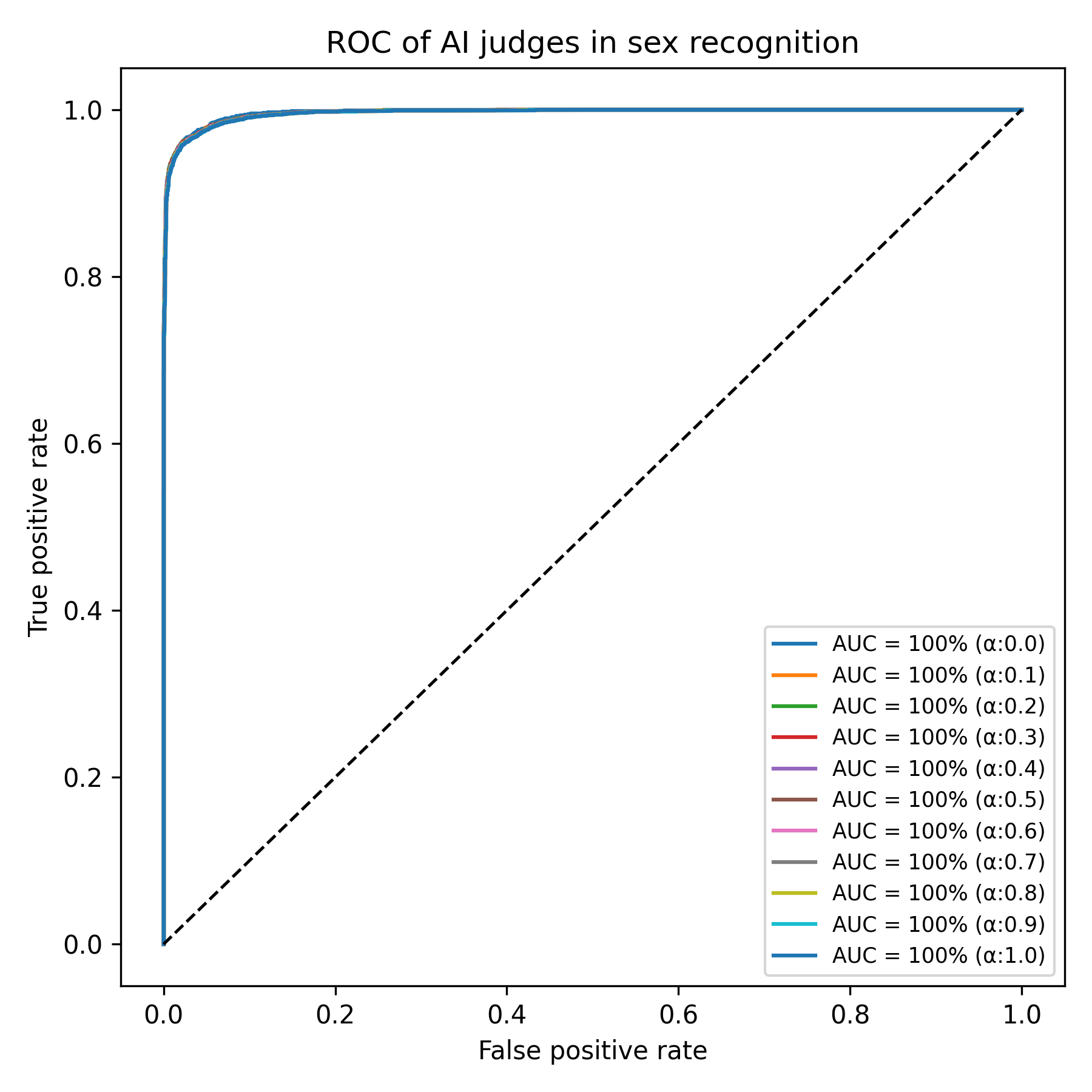}
  \caption{
  ROC curves of the sex-leakage judge on the same CNN-neutralised images.
  All curves lie near the upper-left corner with an AUC of 100 \%,
  showing that the CNN neutraliser leaves sex cues fully intact across the
  entire \(\alpha\) range.
}
  \label{fig:cnn_roc_sex}
\end{figure}

\clearpage
\begin{figure}[htbp]
  \centering
  \includegraphics[width=\textwidth]{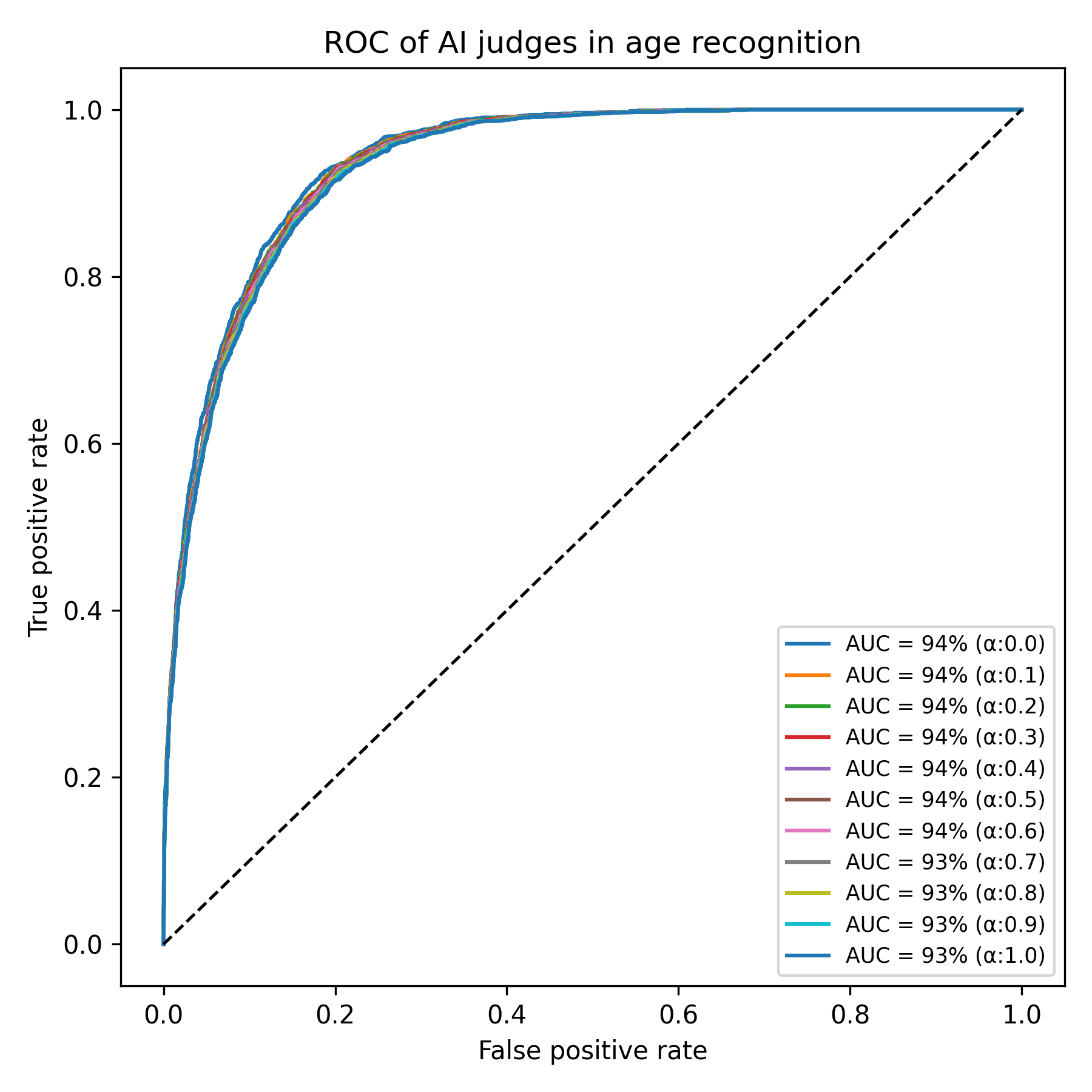}
  \caption{
  ROC curves of the age-leakage judge evaluated on images edited by the
  CNN neutraliser. Line colour encodes the neutralisation weight
  $\alpha \in \{0.0,\dots,1.0\}$, but the curves and AUCs (93--94\%)
  almost coincide, indicating that the CNN edit fails to remove
  age-related cues from the X-rays.
  }
  \label{fig:cnn_roc_age}
\end{figure}

\clearpage
\subsection{Grad-CAM visualisations}\label{app:gradcam}

\begin{figure}[htbp]
  \centering
  \includegraphics[width=0.48\linewidth]{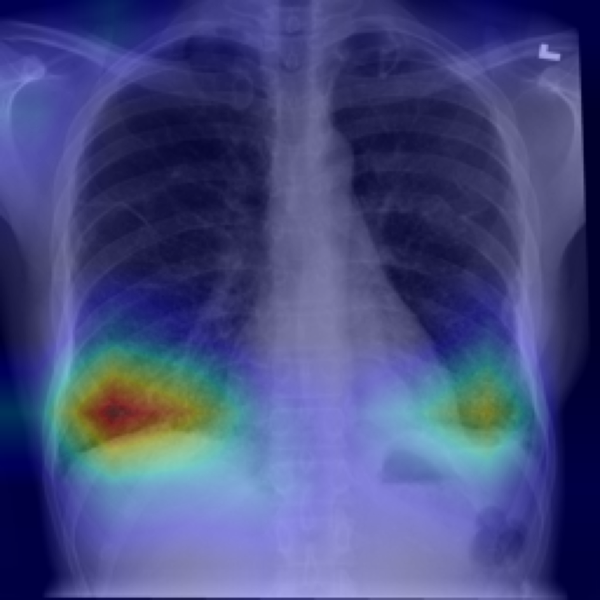}
  \includegraphics[width=0.48\linewidth]{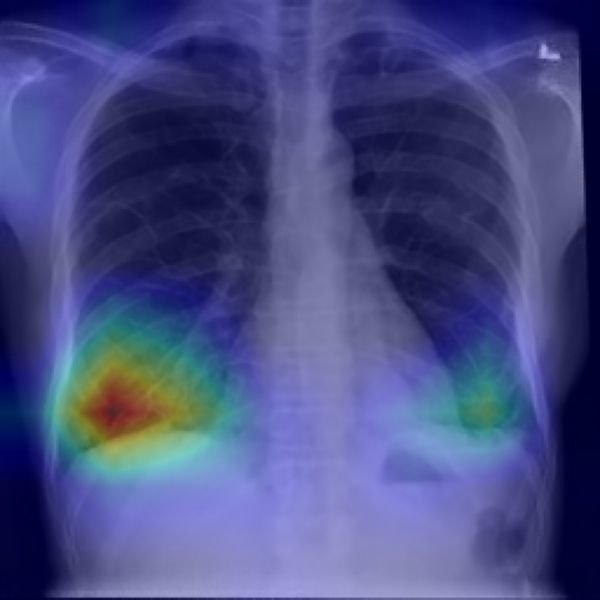}
  \caption{CNN Grad-CAM overlays at $\alpha=0$ and $\alpha=0.5$.}
  \label{fig:gradcam_cnn}
\end{figure}

\begin{figure}[htbp]
  \centering
  \includegraphics[width=0.48\linewidth]{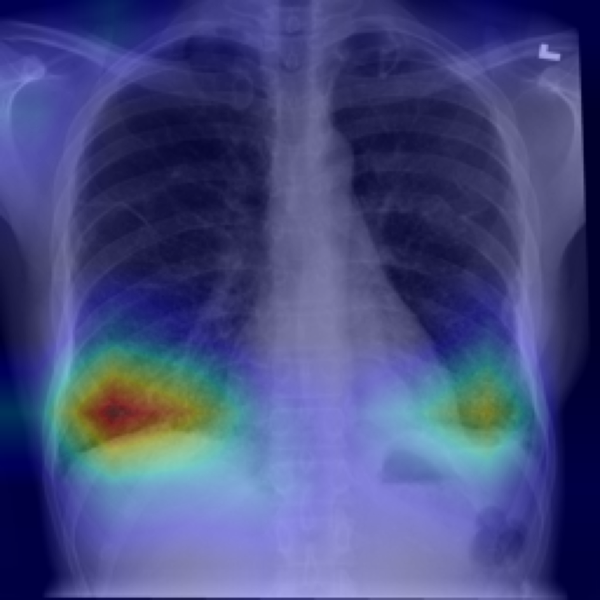}
  \includegraphics[width=0.48\linewidth]{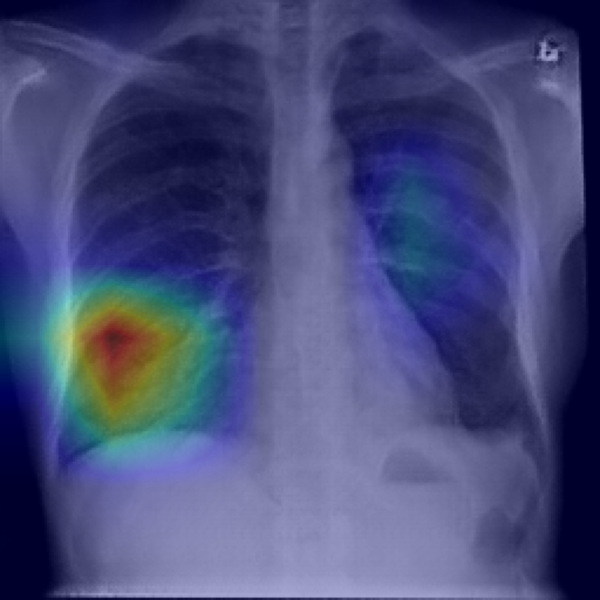}
  \caption{ViT Grad-CAM overlays at $\alpha=0$ and $\alpha=0.5$.}
  \label{fig:gradcam_vit}
\end{figure}

\clearpage
\begin{figure}[t]
  \centering
  \includegraphics[width=\linewidth]{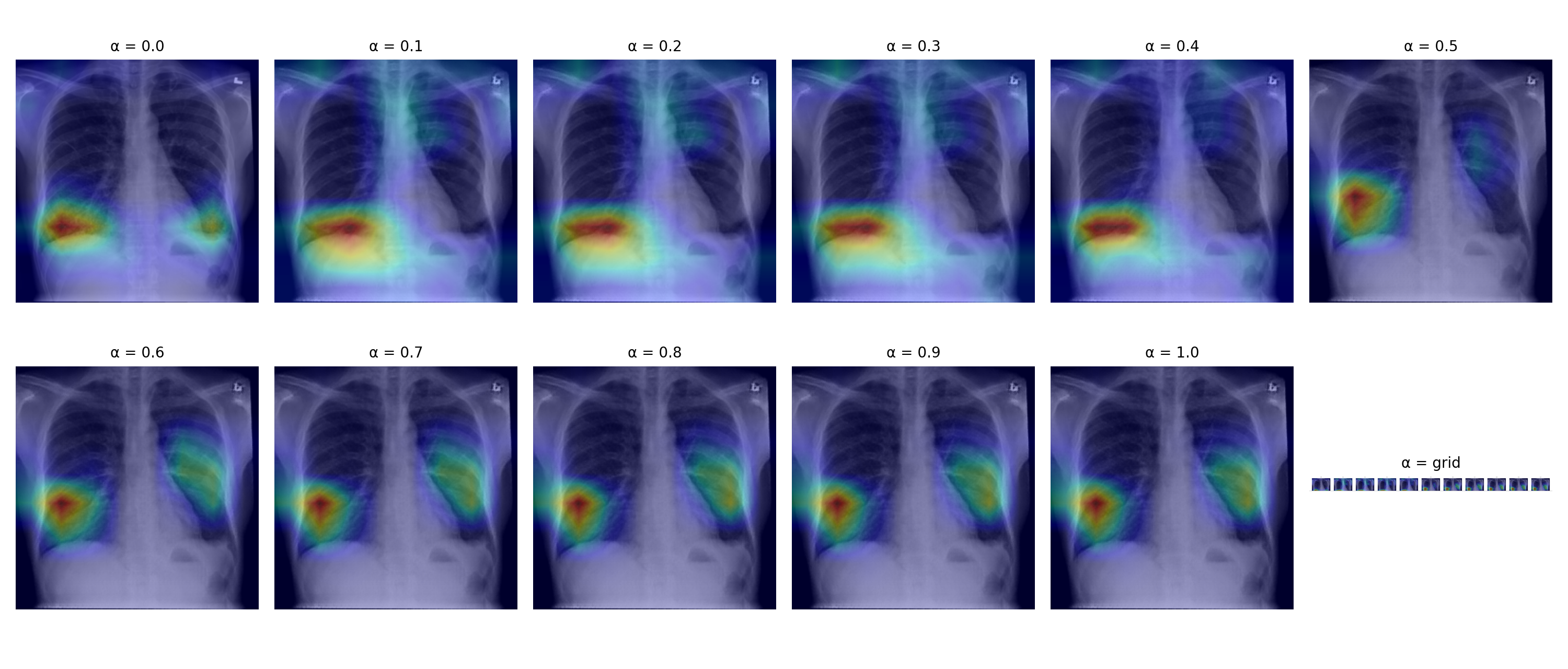}
  \caption{%
    Grad-CAM maps for the gender AI-Judge on the \emph{same} test X-ray after
    progressively stronger ViT–neutralisation
    (\(\alpha = 0.0, 0.1, \dots, 1.0\), left-to-right).
    Warm colours (red / yellow) mark pixels whose feature activations most
    \emph{increase} the “male” logit, while blue regions have little effect.
    Each panel is independently normalised to \([0,1]\); colour intensity is
    therefore comparable only \emph{within} a panel.
    As \(\alpha\) grows, the lower-lung hotspot fades and shrinks—mirrored by
    the model’s output probability for the male class, which drops from
    \(p_{\text{male}}=0.99\) at \(\alpha=0.0\) to \(p_{\text{male}}=0.01\) at
    \(\alpha\ge 0.7\).%
  }
  \label{fig:gradcam_grid}
  \par\footnotesize
  Male–class probabilities for the same image:
  0.993, 0.995, 0.995, 0.996, 0.996, 0.196, 0.009, 0.008, 0.009, 0.009, 0.011.
\end{figure}

\clearpage
\subsection{neutralized plates}\label{app:plates}

\clearpage
\begin{figure}[p]
  \centering
  \includegraphics[width=\textwidth]{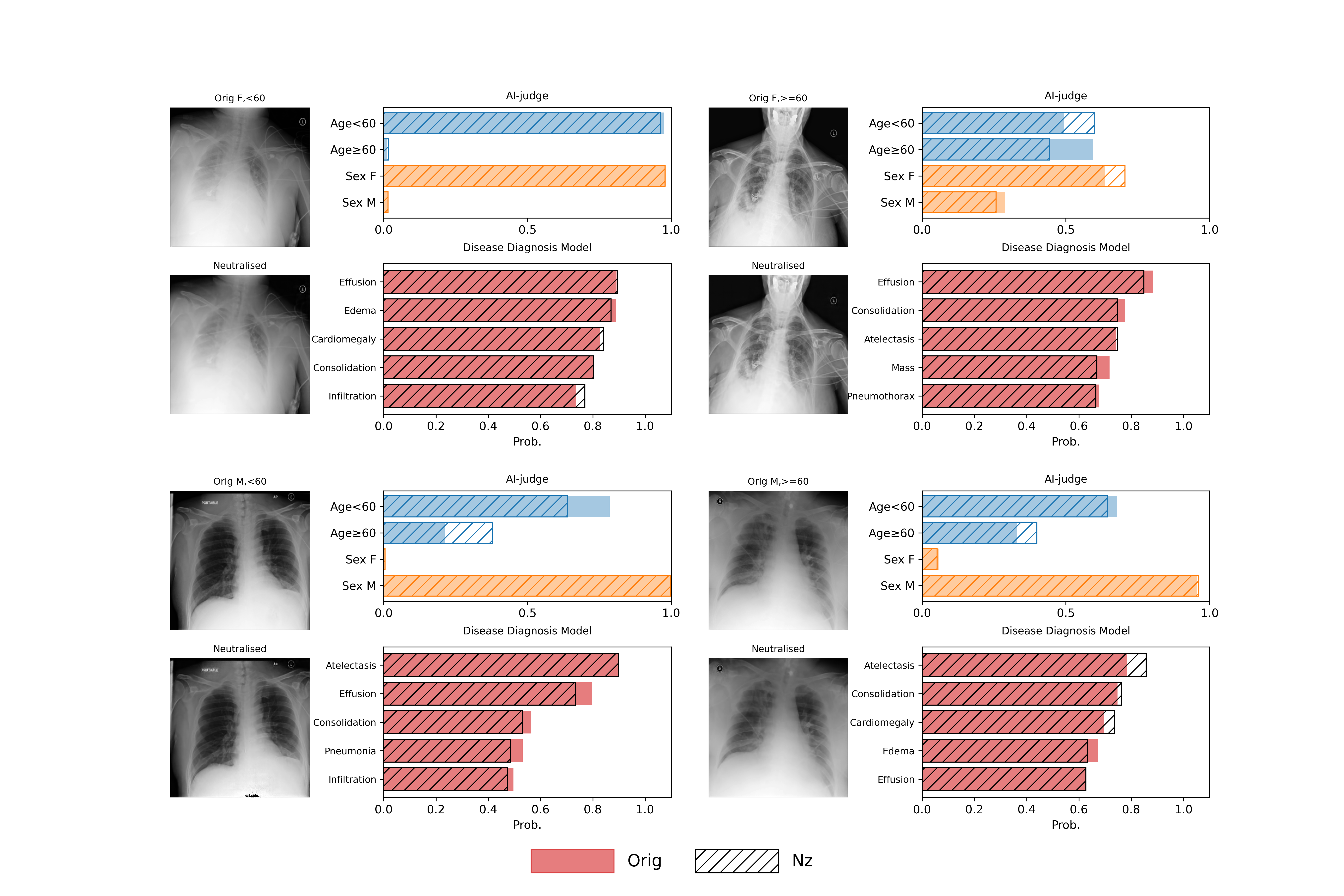}
  \caption{Illustrative plate for the \textbf{CNN} neutralizer at half strength
           ($\alpha = 0.5$). For each subject (female $< 60$\,y, female $\geq 60$\,y,
           male $< 60$\,y, male $\geq 60$\,y), the original CXR (left of each pair) is
           shown beside its neutralized counterpart (right), together with
           AI-Judge outputs (Age, Sex) and the five most probable DDM
           findings. Neutralization drives Age scores towards 0.5 or flips
           the Sex class, but disease probabilities are visibly distorted,
           mirroring the large macro-AUC drop.}
  \label{fig:plate_cnn}
\end{figure}

\clearpage
\begin{figure}[p]
    \centering
    \includegraphics[width=\textwidth]{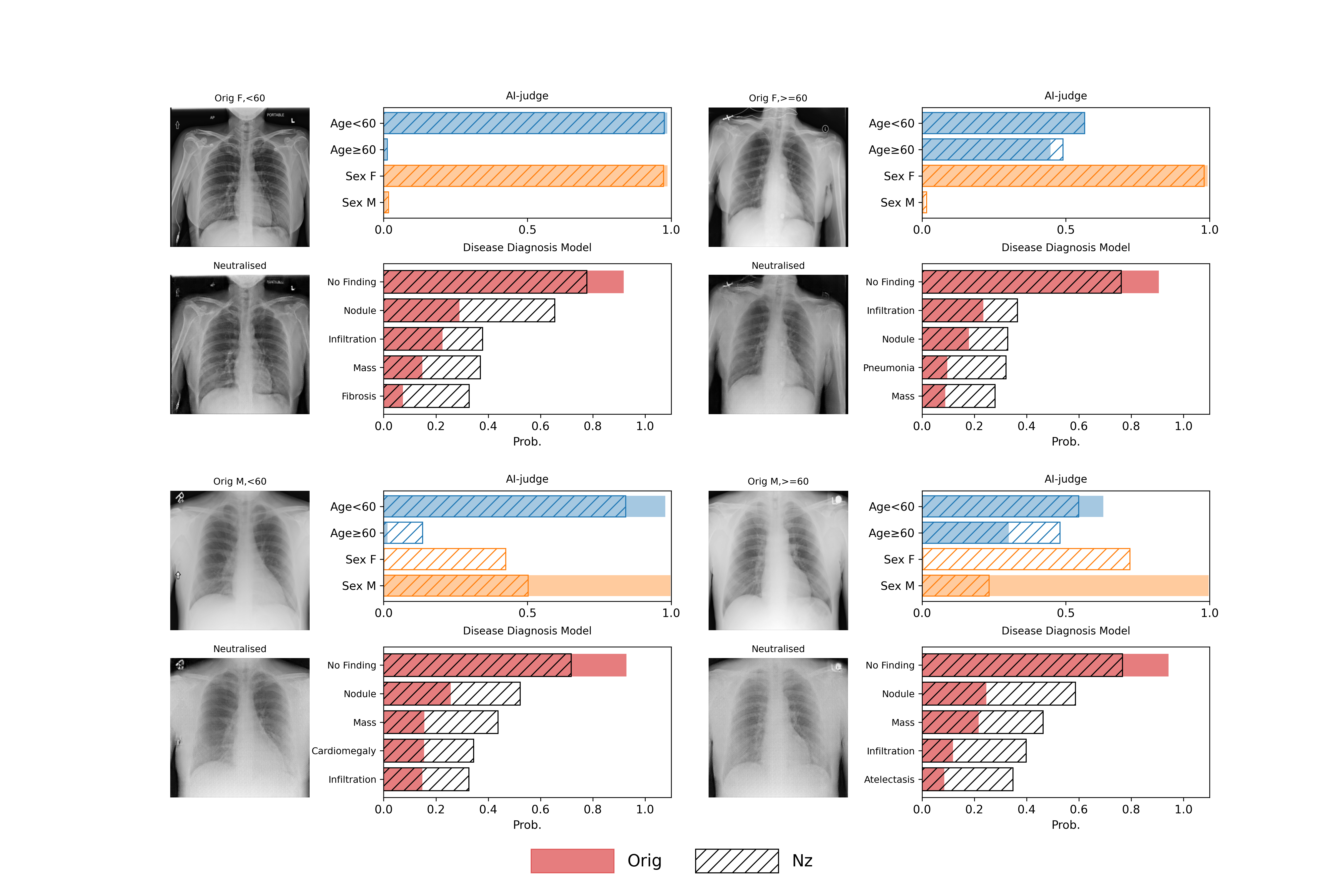}
    \caption{Illustrative plate for the \textbf{ViT} neutralizer at half strength
           ($\alpha=0.5$).  The same four subjects are shown.  AI-Judge
           outputs collapse towards 0.5 (Age) or flip the Sex prediction,
           confirming removal of protected cues, while the DDM probabilities
           remain almost unchanged—demonstrating that ViT-based
           neutralization preserves diagnostic content.}
    \label{fig:plate_vit}
  \end{figure}

\clearpage
\subsection{Critical-difference diagram}\label{app:cd_diagram}
\begin{figure}[htbp]
  \centering
  \includegraphics[width=0.75\linewidth]{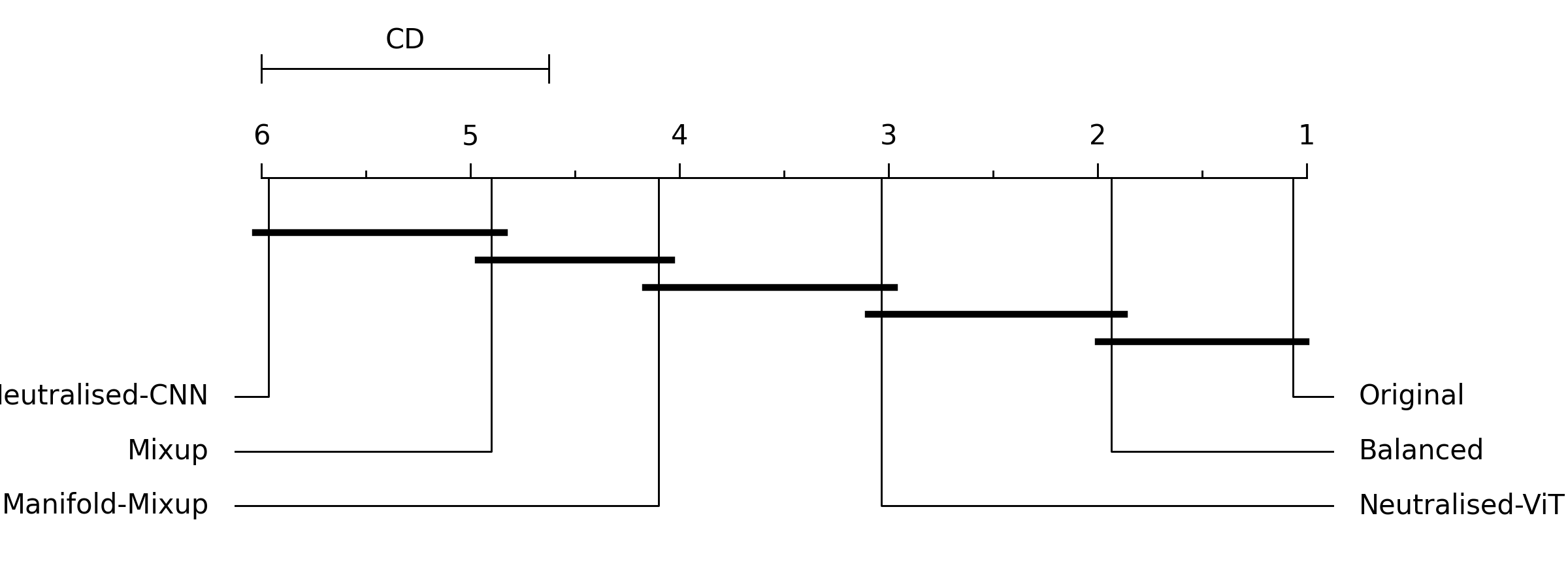}
  \caption{
    Critical-difference diagram (Friedman + Nemenyi, $\alpha = 0.05$)
    comparing six debiasing strategies across 15 findings and two protected
    attributes (30 tasks in total). Smaller average rank indicates better
    ROC-AUC. Methods connected by a thick line do not differ
    significantly. Neutralised-ViT achieves the second-best rank and is not
    significantly worse than the Original baseline, whereas Neutralised-CNN
    is statistically the worst.}
  \label{fig:cd_diag}
\end{figure}

\subsection{Macro-performance table}\label{app:macro}

\begin{table}[ht]
  \centering
  \scriptsize
  \setlength{\tabcolsep}{2pt}
  \caption{Macro-average performance of the disease-diagnosis models (DDMs) on the
    ChestX-ray14 test set. Each entry is the mean over 15 findings, higher is
    better for all metrics. ``Original'' is the baseline ConvNeXt trained on the
    imbalanced data. ``Balanced'' reweights the loader to equalise
    protected-attribute bins. ``Neutralised'' replaces the input with images
    edited by the attribute-neutraliser (either CNN or ViT encoder). Mixup and
    Manifold-Mixup are data-augmentation baselines. The last two rows reproduce
    the CNN-neutralised results reported by Hu \textit{et al.} \cite{hu2024enhancing};
    their AUC is shown for reference only and was not included in the statistical tests.}
  \label{tab:macro_chestx}
  \begin{tabular*}{\columnwidth}{@{\extracolsep{\fill}} llrrrrrr}
    \toprule
    Method                & Attribute & ROC-AUC & ACC   & SEN   & SPE   & $F_{1}$ & PR-AUC \\
    \midrule
    Original              & Age       & 0.802   & 0.751 & 0.723 & 0.753 & 0.232   & 0.234  \\
                          & Sex       & 0.802   & 0.751 & 0.723 & 0.753 & 0.232   & 0.234  \\
    \midrule
    Balanced Sampling     & Age       & 0.792   & 0.709 & 0.748 & 0.709 & 0.217   & 0.212  \\
                          & Sex       & 0.793   & 0.710 & 0.757 & 0.710 & 0.221   & 0.216  \\
    \midrule
    neutralized (CNN)     & Age       & 0.503   & 0.521 & 0.526 & 0.512 & 0.116   & 0.086  \\
                          & Sex       & 0.493   & 0.568 & 0.489 & 0.536 & 0.126   & 0.084  \\
    neutralized (ViT)     & Age       & 0.759   & 0.686 & 0.721 & 0.686 & 0.206   & 0.196  \\
                          & Sex       & 0.777   & 0.709 & 0.730 & 0.711 & 0.216   & 0.206  \\
    \midrule
    Mixup                 & Age       & 0.644   & 0.557 & 0.714 & 0.541 & 0.165   & 0.118  \\
                          & Sex       & 0.642   & 0.591 & 0.659 & 0.584 & 0.166   & 0.125  \\
    Manifold-Mixup        & Age       & 0.690   & 0.633 & 0.691 & 0.629 & 0.182   & 0.154  \\
                          & Sex       & 0.723   & 0.668 & 0.700 & 0.668 & 0.192   & 0.149  \\
    \midrule
    Neutralized (CNN)$^{\star}$ & Age & 0.8057 & 0.8287 & 0.5768 & 0.8337 & 0.2761 & 0.2720 \\
    Neutralized (CNN)$^{\star}$ & Sex & 0.8054 & 0.8413 & 0.5525 & 0.8480 & 0.2874 & 0.2742 \\
    \bottomrule
  \end{tabular*}
  \par\smallskip
  {\footnotesize $^{\star}$Reported from Hu\,\textit{et al.} \cite{hu2024enhancing}, Table~S9.}
\end{table}

\clearpage
\section{Project Scope and Resource Constraints}
\label{app:scope}

This study was originally planned to (i) apply the full Attribute-Neutral Framework to four chest-X-ray datasets (ChestX-ray14, CheXpert, MIMIC-CXR, PadChest), (ii) compare CNN and Vision-Transformer generators, and (iii) evaluate the additional debiasing baselines on all four datasets. Three obstacles forced the scope to be pared down. Each AttGAN training cycle required substantial GPU resources, quickly exhausting the available Snellius quota, missing or unstable image-generation code required extensive debugging, and these delays left too little time to complete runs on all four chest-X-ray datasets.

The complete CNN pipeline on Chest-Xray14 ran first and used most of the Snellius GPU quota, leaving limited resources for the ViT extension and resulting in the ViT neutralizer being trained under a reduced compute budget. Consequently, the end-to-end pipeline was executed only on Chest-X-ray14, the other datasets are covered by conducting Exploratory Data Analysis of patient subgroup demographics and disease distributions (Appendix~\ref{app:eda}). Practical fallout of the down-scaling is detailed in the ViT-training subsection of Methodology and reconsidered in the Limitations.

Several unanticipated challenges arose despite obtaining GPU allocations on the Snellius cluster and acquiring multiple Google Colab subscriptions in an effort to adapt the transformer models:

\begin{itemize}
  \item \textbf{Compute demands.} Each end-to-end training cycle (Neutralizer → AI-Judge → Diagnosis Model) on a large ChestX-ray dataset required an enormous amount of GPU resources, especially after extensive debugging, testing, and multiple full-length runs that produced no errors until final evaluation, exceeding available GPU quota.
  \item \textbf{Code instability.} A major script for the Attribute Neutralizer's image generation was missing, necessitating extensive debugging and re-implementation before any reliable stable training could begin.
  \item \textbf{Timeline pressure.} The long training times required for AttGAN-based models to perform well made it infeasible to complete the planned experimental design across multiple datasets and architectures within the thesis timeframe.
\end{itemize}

As a result, the experimental scope was revised to ensure a complete pipeline within the available resources:
\begin{itemize}
  \item The complete Attribute Neutral Framework pipeline run was conducted on ChestX-ray14, all results refer to this dataset.
  \item For CheXpert, MIMIC-CXR, and PadChest, exploratory data analyses and subgroup data analysis were conducted.
\end{itemize}

This narrowed scope limits external validity but allows the following:
\begin{enumerate}
  \item A complete, end-to-end demonstration of the Attribute Neutral Framework.
  \item Clear documentation of the practical challenges (compute demands, code stability) that future researchers must address when scaling to multiple CXR datasets or architectures.
\end{enumerate}

\end{document}